\documentclass[letterpaper]{article} %

\usepackage{aaai25}  %
\usepackage[table,xcdraw]{xcolor}
\usepackage{times}  %
\usepackage{helvet}  %
\usepackage{courier}  %
\usepackage[hyphens]{url}  %
\usepackage{graphicx} %
\usepackage{natbib}  %
\usepackage{caption} %
\usepackage{algorithm}
\usepackage{algorithmic}
\usepackage{pifont}
\usepackage{enumitem}

\usepackage{subcaption}
\usepackage{booktabs}
\usepackage[para,online,flushleft]{threeparttable}
\usepackage[textsize=tiny]{todonotes}

\usepackage{multirow}
\usepackage{amsfonts}

\usepackage{xspace}
\usepackage{siunitx}

\newcommand{\ftwlong}{Fields of The World\xspace}
\newcommand{\ftw}{\texttt{FTW}\xspace}
\newcommand{\patchtotal}{\num{70462}\xspace}
\newcommand{\totalcountries}{24\xspace}

\usepackage{newfloat}
\usepackage{listings}
\DeclareCaptionStyle{ruled}{labelfont=normalfont,labelsep=colon,strut=off} %
\floatstyle{ruled}
\newfloat{listing}{tb}{lst}{}
\floatname{listing}{Listing}
\title{\ftwlong: A Machine Learning Benchmark Dataset For Global Agricultural Field Boundary Segmentation}
\author{
    Hannah Kerner\textsuperscript{\rm 1}\equalcontrib,
    Snehal Chaudhari\textsuperscript{\rm 1}\equalcontrib,
    Aninda Ghosh\textsuperscript{\rm 1}\equalcontrib,
    Caleb Robinson\textsuperscript{\rm 2}\equalcontrib,\\
    Adeel Ahmad\textsuperscript{\rm 3 4},
    Eddie Choi\textsuperscript{\rm 4},
    Nathan Jacobs\textsuperscript{\rm 4},
    Chris Holmes\textsuperscript{\rm 5},
    Matthias Mohr\textsuperscript{\rm 5},\\
    Rahul Dodhia\textsuperscript{\rm 2},
    Juan M. Lavista Ferres\textsuperscript{\rm 2},
    Jennifer Marcus\textsuperscript{\rm 5}
}
\affiliations{
    \textsuperscript{\rm 1}Arizona State University, Tempe, AZ 85281 USA, hkerner@asu.edu\\

    \textsuperscript{\rm 2}Microsoft AI for Good Research Lab, Redmond, WA 98052 USA \\

    \textsuperscript{\rm 3}Taylor Geospatial Institute, St Louis, MO 63108 USA \\

    \textsuperscript{\rm 4}Washington University in St Louis, St Louis, MO 63130 USA \\

    \textsuperscript{\rm 5}Taylor Geospatial Engine, St Louis, MO 63130 USA \\
}

\begin{document}

\maketitle

\begin{abstract}
Crop field boundaries are foundational datasets for agricultural monitoring and assessments but are expensive to collect manually. 
Machine learning (ML) methods for automatically extracting field boundaries from remotely sensed images could help realize the demand for these datasets at a global scale. However, current ML methods for field instance segmentation lack sufficient geographic coverage, accuracy, and generalization capabilities. 
Further, research on improving ML methods is restricted by the lack of labeled datasets representing the diversity of global agricultural fields. We present \ftwlong (\ftw)---a novel ML benchmark dataset for agricultural field instance segmentation spanning \totalcountries countries on four continents (Europe, Africa, Asia, and South America). \ftw is an order of magnitude larger than previous datasets with \patchtotal samples, each containing instance and semantic segmentation masks paired with multi-date, multi-spectral Sentinel-2 satellite images. We provide results from baseline models for the new \ftw benchmark, show that models trained on \ftw have better zero-shot and fine-tuning performance in held-out countries than models that aren't pre-trained with diverse datasets, and show positive qualitative zero-shot results of \ftw models in a real-world scenario -- running on Sentinel-2 scenes over Ethiopia.
\end{abstract}

\begin{links}
    \link{Code}{https://github.com/fieldsoftheworld/ftw-baselines} %
    \link{Datasets}{https://beta.source.coop/repositories/kerner-lab/fields-of-the-world/} %
\end{links}

\section{Introduction}
\label{sec:intro}

\begin{figure}[ht!]
    \centering
    \includegraphics[width=1\linewidth]{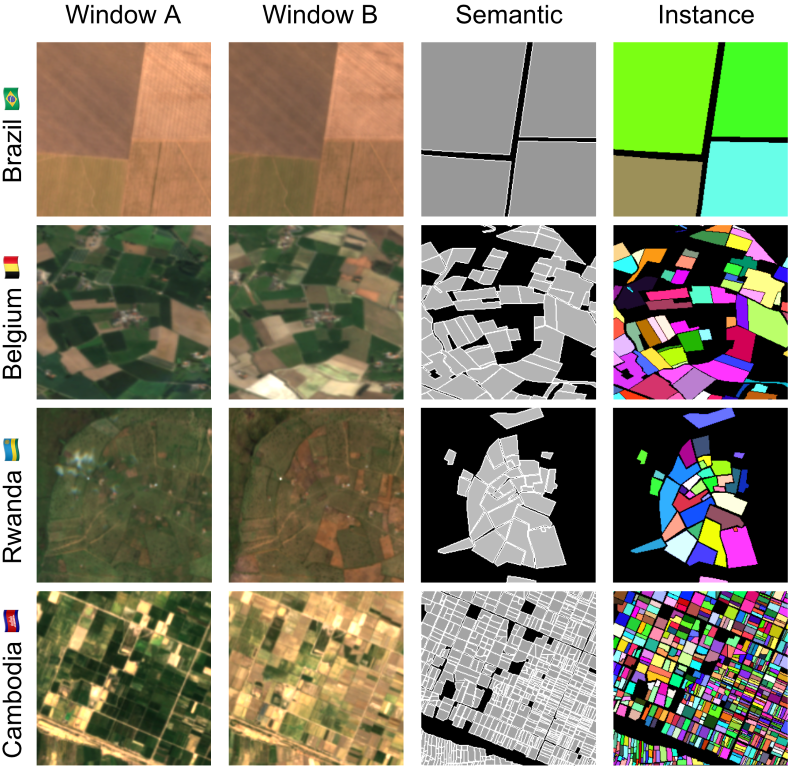}
    \caption{Training samples from four continents, demonstrating the diversity within \ftwlong.}
    \label{fig:teaser}
\end{figure}

Crop field boundary datasets are urgently needed in agricultural monitoring, sustainable agriculture, and development applications~\cite{nakalembe2023-considerations}. However, these datasets do not exist for most of the world. Automatic field delineation in globally available satellite imagery offers a promising solution, but semantic reasoning about globally diverse agricultural landscapes in satellite imagery remains challenging. Field morphologies, agricultural practices, and climate patterns vary greatly across the world. For example, average field sizes range from 1.6 hectares (ha) in Sub-Saharan Africa to 121 ha in North America~\cite{debats2016generalized}. Motivated by this global diversity, we present a novel dataset for agricultural field instance segmentation: \ftwlong (\ftw). \ftw spans diverse landscapes in \totalcountries countries on 4 continents (Europe, Asia, Africa, and South America). \ftwlong aims to catalyze field instance segmentation research and enable consistent, granular evaluation of different modeling approaches.

Field boundary datasets enable field-scale monitoring of crop conditions, yield, pest/diseases, farming practices, resource utilization, and other agricultural characteristics~\cite{nakalembe2023-considerations}. They are also in high demand for conservation and climate change policies and programs that require Measurement, Reporting, and Verification (MRV) of greenhouse gas emissions, carbon sequestration, and sustainable land management practices, such as the European Union Deforestation Regulation~\cite{eu_deforestation_regulation_2023}. Field boundaries also simplify challenging tasks like crop-type classification by enabling classification at the object level rather than over individual pixels~\cite{garnot2022multi}. Statistics agencies use field boundaries for ground-based survey design~\cite{nakalembe2023-considerations}. Field boundary maps over multiple years enable analyses of environmental and socioeconomic land change dynamics such as aggregation or fragmentation of farm parcels over time~\cite{estes2022high,sullivan2023large}.

Previous work has demonstrated good performance for field instance segmentation in European countries, enabled by a combination of novel algorithms and labeled datasets (e.g.,~\citet{wang2022unlocking,garnot2021panoptic}). Research progress has been driven by benchmark datasets such as PASTIS~\cite{garnot2021panoptic,garnot2022multi}, AI4Boundaries~\cite{d2023ai4boundaries}, and AI4FoodSecurity~\cite{ai4foodsecurity}. 
While these datasets have been critical research catalysts, they do not fully capture the diversity and complexity of global agricultural landscapes.
Existing datasets have limited geographic diversity, with labels concentrated in a handful of (usually European) countries (Table~\ref{tab:relatedwork}). 

\begin{figure}[t!]
    \centering
    \includegraphics[width=1\linewidth]{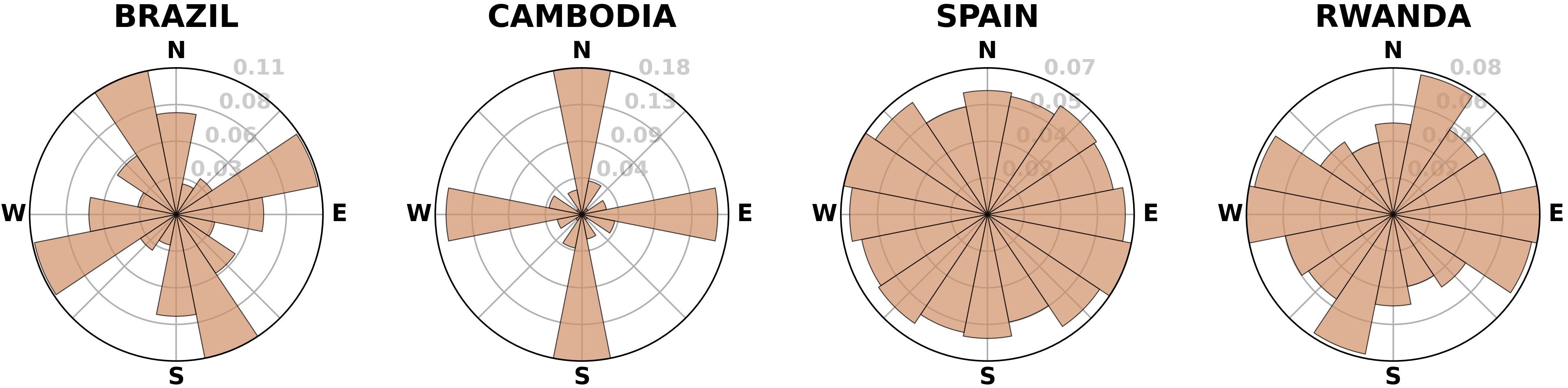}
    \caption{Field orientation histograms for selected countries.} %
    \label{fig:orient}
\end{figure}

\ftwlong captures greater geographic diversity, morphological diversity, and agro-climatic diversity than any previous dataset. It includes fields of various sizes (Figure~\ref{fig:area}), shape, and orientation (Figure~\ref{fig:orient}). \ftw is also an order of magnitude larger than previous datasets, with \patchtotal samples covering a total geographic area of \qty{166293}{\square\km}.

\ftwlong provides harmonized ML-ready inputs from the optical Sentinel-2 satellite. Each example in \ftwlong includes four spectral channels (red, green, blue, and near-infrared) from two contrasting dates. Labels include instance and semantic segmentation masks. We provide the polygon label annotations in a standardized format using the \textit{fiboa} (field boundaries for agriculture) specification~\cite{fiboa-spec}. This makes previously siloed datasets interoperable and enables users to obtain custom satellite data or other inputs corresponding to geo-referenced labels. The provided metadata also allows users to easily subset the dataset depending on their needs (e.g., commercial license or specific location). 

We include training, validation, and test sets for each country, using existing splits when possible to maximize compatibility with existing work. We propose benchmark tasks that mimic real-world scenarios relevant to downstream users of field boundary datasets (e.g., region-specific evaluation, transfer learning, and zero-shot generalization). Finally, we perform experiments to demonstrate the value of the \ftw dataset and provide baseline results for benchmark tasks.
We release code via Github, data via Source Cooperative, and data loaders and pre-trained models via TorchGeo.

\section{Dataset Description}
\label{sec:dataset-description}

\begin{table*}[t]
\caption{Key dataset details for each country in \ftw, with green and brown cells indicating Windows A and B respectively.}
  \label{tab:dataset-details}
  \centering
\resizebox{\textwidth}{!}{%
\begin{tabular}{lllllllllllllcll}
\toprule
\multicolumn{16}{c}{\textbf{Presence/absence labels}} \\
\multicolumn{1}{l}{Country} &
  Jan &
  Feb &
  Mar &
  Apr &
  May &
  Jun &
  Jul &
  Aug &
  Sep &
  Oct &
  Nov &
  Dec &
  Sub-sampled? &
  \# train, val, test &
  Source \\
  \midrule
\multirow{2}{*}{Austria} &
   &
   &
   &
  \cellcolor[HTML]{6AA84F} &
  \cellcolor[HTML]{6AA84F} &
  \cellcolor[HTML]{6AA84F} &
  \cellcolor[HTML]{6AA84F} &
   &
   &
   &
   &
   & \multirow{2}{*}{\checkmark}
   & \multirow{2}{*}{5303, 637, 745}
   & \multirow{2}{*}{\citet{schneider2022eurocrops21}}
   \\
 &
   &
   &
   &
   &
   &
   &
  \cellcolor[HTML]{BF9000} &
  \cellcolor[HTML]{BF9000} &
  \cellcolor[HTML]{BF9000} &
  \cellcolor[HTML]{BF9000} &
   &
   \\ \hline
 &
   &
   &
   &
   &
   &
   &
  \cellcolor[HTML]{6AA84F} &
  \cellcolor[HTML]{6AA84F} &
  \cellcolor[HTML]{6AA84F} &
  \cellcolor[HTML]{6AA84F} &
   &
   & \multirow{2}{*}{\checkmark}
   & \multirow{2}{*}{1554, 189, 198}
   & \multirow{2}{*}{\citet{schneider2022eurocrops21}}
   \\ 
\multirow{-2}{*}{Belgium} &
   &
  \cellcolor[HTML]{BF9000} &
  \cellcolor[HTML]{BF9000} &
  \cellcolor[HTML]{BF9000} &
  \cellcolor[HTML]{BF9000} &
  \cellcolor[HTML]{BF9000} &
   &
   &
   &
   &
   &
   \\ \hline
    &
   &
   &
   &
   &
  \cellcolor[HTML]{6AA84F} &
  \cellcolor[HTML]{6AA84F} &
  \cellcolor[HTML]{6AA84F} &
  \cellcolor[HTML]{6AA84F} &
  \cellcolor[HTML]{6AA84F} &
  \cellcolor[HTML]{6AA84F} &
  \cellcolor[HTML]{6AA84F} &
       & \multirow{2}{*}{\ding{55}}
       & \multirow{2}{*}{245, 27, 25}
   & \multirow{2}{*}{\citet{persello2023ai4smallfarms}}
   \\
\multirow{-2}{*}{Cambodia} &
   &
   &
  \cellcolor[HTML]{BF9000} &
  \cellcolor[HTML]{BF9000} &
  \cellcolor[HTML]{BF9000} &
  \cellcolor[HTML]{BF9000} &
  \cellcolor[HTML]{BF9000} &
  \cellcolor[HTML]{BF9000} &
  \cellcolor[HTML]{BF9000} &
   &
  \cellcolor[HTML]{BF9000} &
  \cellcolor[HTML]{BF9000} 
  \\ \hline
 &
   &
   &
   &
  \cellcolor[HTML]{6AA84F} &
  \cellcolor[HTML]{6AA84F} &
  \cellcolor[HTML]{6AA84F} &
  \cellcolor[HTML]{6AA84F} &
  \cellcolor[HTML]{6AA84F} &
  \cellcolor[HTML]{6AA84F} &
  \cellcolor[HTML]{6AA84F} &
   &
   & \multirow{2}{*}{\checkmark}
   & \multirow{2}{*}{2778, 351, 353}
   & \multirow{2}{*}{\citet{arkod}}
   \\
\multirow{-2}{*}{Croatia} &
   &
   &
   &
   &
  \cellcolor[HTML]{BF9000} &
  \cellcolor[HTML]{BF9000} &
  \cellcolor[HTML]{BF9000} &
   &
  \cellcolor[HTML]{BF9000} &
  \cellcolor[HTML]{BF9000} &
   &
   \\ \hline
 &
   &
   &
   &
   &
   &
   &
  \cellcolor[HTML]{6AA84F} &
  \cellcolor[HTML]{6AA84F} &
  \cellcolor[HTML]{6AA84F} &
  \cellcolor[HTML]{6AA84F} &
   &
   & \multirow{2}{*}{\checkmark}
   & \multirow{2}{*}{2868, 360, 332}
   & \multirow{2}{*}{\citet{landbrugsGIS}}
   \\
\multirow{-2}{*}{Denmark} &
  \cellcolor[HTML]{BF9000} &
  \cellcolor[HTML]{BF9000} &
  \cellcolor[HTML]{BF9000} &
  \cellcolor[HTML]{BF9000} &
  \cellcolor[HTML]{BF9000} &
  \cellcolor[HTML]{BF9000} &
   &
   &
   &
   &
   &
   \\ \hline
 &
   &
   &
   &
  \cellcolor[HTML]{6AA84F} &
  \cellcolor[HTML]{6AA84F} &
  \cellcolor[HTML]{6AA84F} &
   &
   &
   &
   &
   &
   & \multirow{2}{*}{\checkmark}
   & \multirow{2}{*}{5348, 681, 684}
   & \multirow{2}{*}{\citet{schneider2022eurocrops21}}
   \\
\multirow{-2}{*}{Estonia} &
   &
   &
   &
   &
   &
   &
  \cellcolor[HTML]{BF9000} &
  \cellcolor[HTML]{BF9000} &
  \cellcolor[HTML]{BF9000} &
   &
   &
   \\ \hline
 &
   &
   &
   &
   &
  \cellcolor[HTML]{6AA84F} &
  \cellcolor[HTML]{6AA84F} &
  \cellcolor[HTML]{6AA84F} &
   &
   &
   &
   &
   & \multirow{2}{*}{\checkmark}
   & \multirow{2}{*}{4527, 550, 588}
   & \multirow{2}{*}{\citet{ruokavirasto}}
   \\
\multirow{-2}{*}{Finland} &
   &
   &
   &
   &
   &
   &
  \cellcolor[HTML]{BF9000} &
  \cellcolor[HTML]{BF9000} &
  \cellcolor[HTML]{BF9000} &
   &
   &
   \\ \hline
 &
   &
   &
   &
   &
   &
  \cellcolor[HTML]{6AA84F} &
  \cellcolor[HTML]{6AA84F} &
  \cellcolor[HTML]{6AA84F} &
  \cellcolor[HTML]{6AA84F} &
  \cellcolor[HTML]{6AA84F} &
   &
   & \multirow{2}{*}{\checkmark}
   & \multirow{2}{*}{1974, 240, 258}
   & \multirow{2}{*}{\citet{rpg}}
   \\
\multirow{-2}{*}{Corsica} &
   &
   &
   &
  \cellcolor[HTML]{BF9000} &
  \cellcolor[HTML]{BF9000} &
  \cellcolor[HTML]{BF9000} &
  \cellcolor[HTML]{BF9000} &
   &
   &
   &
   &
   \\ \hline
 &
   &
   &
   &
  \cellcolor[HTML]{6AA84F} &
  \cellcolor[HTML]{6AA84F} &
  \cellcolor[HTML]{6AA84F} &
   &
   &
   &
   &
   &
   & \multirow{2}{*}{\checkmark}
   & \multirow{2}{*}{2773, 339, 396}
   & \multirow{2}{*}{\citet{rpg}}
   \\
\multirow{-2}{*}{France} &
   &
   &
   &
   &
   &
   &
   &
  \cellcolor[HTML]{BF9000} &
  \cellcolor[HTML]{BF9000} &
   &
   &
   \\ \hline
 &
   &
   &
   &
   &
   &
   &
  \cellcolor[HTML]{6AA84F} &
  \cellcolor[HTML]{6AA84F} &
  \cellcolor[HTML]{6AA84F} &
   &
   &
   & \multirow{2}{*}{\ding{55}}
   & \multirow{2}{*}{306, 30, 350}
   & \multirow{2}{*}{\citet{kondmann2021denethor}}
   \\
\multirow{-2}{*}{Germany} &
   &
   &
   &
  \cellcolor[HTML]{BF9000} &
  \cellcolor[HTML]{BF9000} &
   &
   &
   &
   &
   &
   &
   \\ \hline
 &
   &
   &
   &
   &
  \cellcolor[HTML]{6AA84F} &
  \cellcolor[HTML]{6AA84F} &
   &
   &
   &
   &
   &
   & \multirow{2}{*}{\checkmark}
   & \multirow{2}{*}{5529, 668, 741}
   & \multirow{2}{*}{\citet{schneider2022eurocrops21}}
   \\
\multirow{-2}{*}{Latvia} &
   &
   &
   &
   &
   &
   &
   &
   &
  \cellcolor[HTML]{BF9000} &
  \cellcolor[HTML]{BF9000} &
   &
   \\ \hline
 &
   &
   &
   &
  \cellcolor[HTML]{6AA84F} &
  \cellcolor[HTML]{6AA84F} &
  \cellcolor[HTML]{6AA84F} &
   &
   &
   &
   &
   &
   & \multirow{2}{*}{\checkmark}
   & \multirow{2}{*}{4208, 522, 528}
   & \multirow{2}{*}{\citet{schneider2022eurocrops21}}
   \\
\multirow{-2}{*}{Lithuania} &
   &
   &
   &
   &
   &
   &
  \cellcolor[HTML]{BF9000} &
  \cellcolor[HTML]{BF9000} &
  \cellcolor[HTML]{BF9000} &
   &
   &
   \\ \hline
 &
   &
   &
   &
   &
   &
   &
   &
   &
  \cellcolor[HTML]{6AA84F} &
  \cellcolor[HTML]{6AA84F} &
  \cellcolor[HTML]{6AA84F} &
   & \multirow{2}{*}{\checkmark}
   & \multirow{2}{*}{643, 81, 84}
   & \multirow{2}{*}{\citet{flik}}
   \\
\multirow{-2}{*}{Luxembourg} &
   &
   &
   &
   &
   &
  \cellcolor[HTML]{BF9000} &
  \cellcolor[HTML]{BF9000} &
  \cellcolor[HTML]{BF9000} &
   &
   &
   &
   \\ \hline
 &
   &
   &
  \cellcolor[HTML]{6AA84F} &
  \cellcolor[HTML]{6AA84F} &
  \cellcolor[HTML]{6AA84F} &
  \cellcolor[HTML]{6AA84F} &
   &
   &
   &
   &
   &
   & \multirow{2}{*}{\checkmark}
   & \multirow{2}{*}{3110, 381, 388}
   & \multirow{2}{*}{\citet{pdok}}
   \\
\multirow{-2}{*}{Netherlands} &
   &
   &
   &
   &
   &
   &
  \cellcolor[HTML]{BF9000} &
  \cellcolor[HTML]{BF9000} &
  \cellcolor[HTML]{BF9000} &
  \cellcolor[HTML]{BF9000} &
  \cellcolor[HTML]{BF9000} &
   \\ \hline
 &
   &
   &
  \cellcolor[HTML]{6AA84F} &
  \cellcolor[HTML]{6AA84F} &
  \cellcolor[HTML]{6AA84F} &
  \cellcolor[HTML]{6AA84F} &
   &
   &
   &
   &
   &
   & \multirow{2}{*}{\checkmark}
   & \multirow{2}{*}{47, 9, 10}
   & \multirow{2}{*}{\citet{ifap2024}}
   \\
\multirow{-2}{*}{Portugal} &
   &
   &
   &
   &
   &
  \cellcolor[HTML]{BF9000} &
  \cellcolor[HTML]{BF9000} &
  \cellcolor[HTML]{BF9000} &
  \cellcolor[HTML]{BF9000} &
  \cellcolor[HTML]{BF9000} &
  \cellcolor[HTML]{BF9000} &
   \\ \hline
 &
   &
   &
  \cellcolor[HTML]{6AA84F} &
  \cellcolor[HTML]{6AA84F} &
  \cellcolor[HTML]{6AA84F} &
  \cellcolor[HTML]{6AA84F} &
   &
   &
   &
   &
   &
   & \multirow{2}{*}{\checkmark}
   & \multirow{2}{*}{3275, 390, 408}
   & \multirow{2}{*}{\citet{slovensko2024}}
   \\
\multirow{-2}{*}{Slovakia} &
   &
   &
   &
   &
   &
   &
  \cellcolor[HTML]{BF9000} &
  \cellcolor[HTML]{BF9000} &
  \cellcolor[HTML]{BF9000} &
  \cellcolor[HTML]{BF9000} &
  \cellcolor[HTML]{BF9000} &
   \\ \hline
 &
   &
   &
   &
   &
  \cellcolor[HTML]{6AA84F} &
  \cellcolor[HTML]{6AA84F} &
  \cellcolor[HTML]{6AA84F} &
  \cellcolor[HTML]{6AA84F} &
   &
   &
   &
   & \multirow{2}{*}{\checkmark}
   & \multirow{2}{*}{1733, 216, 228}
   & \multirow{2}{*}{\citet{schneider2022eurocrops21}}
   \\
\multirow{-2}{*}{Slovenia} &
   &
   &
   &
   &
   &
   &
   &
   &
  \cellcolor[HTML]{BF9000} &
  \cellcolor[HTML]{BF9000} &
   &
   \\ \hline
 &
   &
   &
   &
   &
   &
   &
  \cellcolor[HTML]{6AA84F} &
  \cellcolor[HTML]{6AA84F} &
  \cellcolor[HTML]{6AA84F} &
   &
   &
   & \multirow{2}{*}{\ding{55}}
   & \multirow{2}{*}{590, 72, 85}
   & \multirow{2}{*}{\citet{planet2021}}
   \\
\multirow{-2}{*}{South Africa} &
   &
   &
   &
  \cellcolor[HTML]{BF9000} &
  \cellcolor[HTML]{BF9000} &
  \cellcolor[HTML]{BF9000} &
   &
   &
   &
   &
   &
   \\ \hline
 &
   &
   &
   &
   &
   &
   &
  \cellcolor[HTML]{6AA84F} &
  \cellcolor[HTML]{6AA84F} &
  \cellcolor[HTML]{6AA84F} &
  \cellcolor[HTML]{6AA84F} &
   &
   & \multirow{2}{*}{\checkmark}
   & \multirow{2}{*}{2015, 201, 216}
   & \multirow{2}{*}{\citet{schneider2022eurocrops21}}
   \\
\multirow{-2}{*}{Spain} &
   &
   &
   &
  \cellcolor[HTML]{BF9000} &
  \cellcolor[HTML]{BF9000} &
  \cellcolor[HTML]{BF9000} &
  \cellcolor[HTML]{BF9000} &
   &
   &
   &
   &
   \\ \hline
 &
   &
   &
   &
  \cellcolor[HTML]{6AA84F} &
  \cellcolor[HTML]{6AA84F} &
  \cellcolor[HTML]{6AA84F} &
   &
   &
   &
   &
   &
   & \multirow{2}{*}{\checkmark}
   & \multirow{2}{*}{3802, 442, 516}
   & \multirow{2}{*}{\citet{blockdataset}}
   \\
\multirow{-2}{*}{Sweden} &
   &
   &
   &
   &
   &
   &
  \cellcolor[HTML]{BF9000} &
  \cellcolor[HTML]{BF9000} &
  \cellcolor[HTML]{BF9000} &
  \cellcolor[HTML]{BF9000} &
   &
      \\ \hline
 &
   &
   &
   &
   &
   &
   &
  \cellcolor[HTML]{6AA84F} &
  \cellcolor[HTML]{6AA84F} &
  \cellcolor[HTML]{6AA84F} &
   &
   &
          & \multirow{2}{*}{\ding{55}}
          & \multirow{2}{*}{228, 36, 23}
   & \multirow{2}{*}{\citet{persello2023ai4smallfarms}}
   \\ 
\multirow{-2}{*}{Vietnam} &
   &
   &
   &
   &
  \cellcolor[HTML]{BF9000} &
  \cellcolor[HTML]{BF9000} &
   &
   &
   &
   &
   &
   \\ 
\midrule
\multicolumn{16}{c}{\textbf{Presence-only labels}} \\
\multicolumn{1}{l}{Country} &
  Jan &
  Feb &
  Mar &
  Apr &
  May &
  Jun &
  Jul &
  Aug &
  Sep &
  Oct &
  Nov &
  Dec &
  Sub-sampled? &
  \# train, val, test &
  Source \\
  \midrule
\multirow{2}{*}{Brazil} &
  \cellcolor[HTML]{6AA84F} &
  \cellcolor[HTML]{6AA84F} &
   &
   &
   &
   &
   &
   &
   &
  \cellcolor[HTML]{6AA84F} &
  \cellcolor[HTML]{6AA84F} &
  \cellcolor[HTML]{6AA84F} 
     & \multirow{2}{*}{\ding{55}}
     & \multirow{2}{*}{1289, 130, 188}
   & \multirow{2}{*}{\citet{oldoni2020}}
  \\
 &
   &
   &
   &
   &
   &
  \cellcolor[HTML]{BF9000} &
  \cellcolor[HTML]{BF9000} &
  \cellcolor[HTML]{BF9000} &
  \cellcolor[HTML]{BF9000} &
   &
   &
   \\ \hline
 &
   &
   &
   &
   &
   &
   &
  \cellcolor[HTML]{6AA84F} &
  \cellcolor[HTML]{6AA84F} &
  \cellcolor[HTML]{6AA84F} &
  \cellcolor[HTML]{6AA84F} &
  \cellcolor[HTML]{6AA84F} &
            & \multirow{2}{*}{\ding{55}}
            & \multirow{2}{*}{1261, 300, 399}
   & \multirow{2}{*}{\citet{Wang_Waldner_Lobell_2023}}
   \\
\multirow{-2}{*}{India} &
   &
   &
  \cellcolor[HTML]{BF9000} &
  \cellcolor[HTML]{BF9000} &
  \cellcolor[HTML]{BF9000} &
  \cellcolor[HTML]{BF9000} &
   &
   &
   &
   &
   &
   \\ \hline
 &
   &
   &
   &
  \cellcolor[HTML]{6AA84F} &
  \cellcolor[HTML]{6AA84F} &
  \cellcolor[HTML]{6AA84F} &
  \cellcolor[HTML]{6AA84F} &
   &
   &
   &
   &
               & \multirow{2}{*}{\ding{55}}
               & \multirow{2}{*}{316, 20, 55}
   & \multirow{2}{*}{\citet{kenya-ecaas}}
   \\
\multirow{-2}{*}{Kenya} &
   &
   &
   &
   &
   &
   &
   &
   &
  \cellcolor[HTML]{BF9000} &
  \cellcolor[HTML]{BF9000} &
  \cellcolor[HTML]{BF9000} &
  \cellcolor[HTML]{BF9000} 
  \\ \hline
 &
   &
   &
   &
  \cellcolor[HTML]{6AA84F} &
  \cellcolor[HTML]{6AA84F} &
   &
   &
   &
   &
   &
   &
               & \multirow{2}{*}{\ding{55}}
               & \multirow{2}{*}{57, 6, 7}
   & \multirow{2}{*}{\citet{nasaharvest2024}}
   \\
\multirow{-2}{*}{Rwanda} &
   &
   &
   &
   &
   &
  \cellcolor[HTML]{BF9000} &
  \cellcolor[HTML]{BF9000} &
  \cellcolor[HTML]{BF9000} &
  \cellcolor[HTML]{BF9000} &
   &
   &
 \\ 
  \bottomrule
\end{tabular}%
}
\end{table*}

\subsection{Annotations}
\subsubsection{Field boundary representations}
Field boundary annotations are typically in the form of geo-referenced polygons. Since field boundaries at the same location may change across growing seasons, these polygons should also be temporally referenced to specify when the boundary is valid. Field polygons may be farmer-reported, manually drawn on high-resolution satellite images with GIS software, or recorded by walking the field perimeter with a handheld location-recording device.
Polygons can then be paired with satellite data from the same location and time.

We conducted a comprehensive search for field polygons from government databases, published literature, and other websites.
We looked for datasets with diverse geographic coverage, high-quality and trustworthy polygon annotations, and licenses that permit reuse. We included all datasets meeting these criteria in \ftw. We considered author-reported quality assessment in each dataset's documentation, previous use of the dataset in ML analyses, and visual inspection (e.g., closed polygons, polygons consistent with satellite images from reported dates, etc). 
Table~\ref{tab:dataset-details} lists the \totalcountries source datasets selected for \ftwlong.

\subsubsection{Presence/absence labels}
\textit{Presence/absence labels} are binary labels providing information about both the occurrence and non-occurrence of a phenomenon across sampled locations or time periods---for example, the presence of a field boundary or its absence. Most of the datasets in \ftwlong have presence/absence labels. However, some have \textit{presence-only} labels, i.e., they are partially labeled. These indicate the presence of some, but not necessarily all, field boundaries in the sampled locations.
Some pixels in presence-only label masks have unknown labels that might be labeled as background.
Partial labels are a common challenge in field boundary segmentation~\cite{wang2022unlocking}. The Rwanda example in Figure~\ref{fig:teaser} (row 3) illustrates presence-only labels while the Cambodia example (row 4) illustrates presence/absence labels.

\subsubsection{Semantic filtering}
We focused \ftwlong on field boundaries for annual crops. Annual crops, also called temporary crops, are planted, grown, and harvested within a single growing season or year. %
Common examples include wheat, rice, maize, soybeans, and barley. This does not include permanent or perennial crops, which are cultivated for longer than one year and are not replanted annually, such as fruit trees, nut trees, and some grasses. We also excluded parcels used for pasture, fallow land, or other non-crop agricultural activities, such as grazing, orchards, vineyards, and forestry. If a dataset included parcels that were not active annual crops, we filtered them out (details in supplement). %

\subsubsection{Sample grids}
Many datasets in Table~\ref{tab:dataset-details}, particularly those sourced from the European Union government websites and EuroCrops~\cite{schneider2022eurocrops21}, have millions of dense annotations spanning the entire country. Including all of these annotations in \ftwlong would bias the dataset toward large European countries. We sub-sampled these datasets by: 1) creating a bounding box enclosing the entire dataset, 2) splitting the bounding box into a grid where each grid cell covered an area between 3300 to 5000 \unit{\square\km}, and 
3) selecting 2-4 grid cells per country that captured a mixture of high-density and low-density agricultural areas.

For the eight datasets in Table~\ref{tab:dataset-details} that were not sub-sampled, we used all field boundaries provided by the source dataset. 
Four of these datasets were published with predefined grids, which we used without modification: Germany~\cite{kondmann2021denethor}, South Africa~\cite{planet2021}, Cambodia~\cite{persello2023ai4smallfarms}, and Vietnam~\cite{persello2023ai4smallfarms}.
The Kenya~\cite{kenya-ecaas} and Brazil~\cite{oldoni2020} label polygons were highly clustered but did not have predefined grids or clusters. We used $k$-means clustering to cluster the label polygons (using the center latitude/longitude of each polygon as the features). We defined a rectangular grid spanning the bounds of each cluster. We chose $k$ by visually inspecting each dataset and balancing between over-clustering (resulting in high overlap between cluster grids) or under-clustering (resulting in sparse grids with large unlabeled areas).
In the India~\cite{Wang_Waldner_Lobell_2023} and Rwanda~\cite{nasaharvest2024} datasets, polygon labels were in small clusters (e.g., 5 fields per cluster for India). We did not define grids for these datasets because we created sample chips directly from each small cluster. 

\subsubsection{Sample chip ROIs}
We tiled each sample grid into $1536\text{m} \times 1536\text{m}$  sample patch ROIs (regions of interest), which we call `chips'.
For India and Rwanda, we created a  $1536\text{m} \times 1536\text{m}$ chip around the center of each label cluster.

\subsubsection{Metadata standardization}
We converted the label polygon datasets to the fiboa specification~\cite{fiboa-spec}. If a dataset was sub-sampled from a larger dataset, we only converted the sub-sampled version. Per the fiboa core specification, each dataset has per-polygon attributes \texttt{id} (unique field identifier), \texttt{determination\_datetime} (last timestamp at which the field boundary existed/was observed), \texttt{area} (field area in hectares), and \texttt{geometry} (field polygon geometry; we use the WGS84/EPSG:4326 coordinate reference system). We included crop type or other attributes when available. Converted GeoParquet 
files are available on Source Cooperative at \url{https://beta.source.coop/repositories/kerner-lab/fields-of-the-world/}. The README for each dataset provides details including the source dataset license (extends to the \ftw subset using it) and link.
We provide GeoParquet files for the sample grids and chip ROIs. %

\subsubsection{Label masks}
Previous work explored several approaches to convert (``rasterize'') field label polygons to label masks, which are the ML prediction targets. Approaches include binary field extent masks (field interiors vs background), binary field boundary masks (field boundaries vs background), 3-class masks (field interiors, field boundaries, and background)~\cite{taravat2021advanced}, and distance masks (to field centroids)~\cite{d2023ai4boundaries}. We provide binary field extent and 3-class semantic masks and instance masks.

\subsection{Satellite data}
\label{sec:satellite-data}
We obtained multispectral Sentinel-2 satellite images using Microsoft Planetary Computer~\cite{microsoft_open_source_2022_7261897}. Images in this catalog are processed to Level 2A (bottom-of-atmosphere) %
and stored in cloud-optimized GeoTIFF (COG) format. We used the red (B04), green (B03), blue (B02), and near-infrared (B08) spectral bands, all of which have spatial resolution of 10 \unit{\m} per pixel. 
We used Sentinel-2  because it is the highest-resolution optical satellite dataset that is freely accessible. %
In the \ftw Github repository, we provide a CSV file containing the Sentinel-2 scene ID, cloud percentage, and date ranges for each sample. %

\subsubsection{Dates}
Previous work showed that contrasting images from different times during the same year improved crop field segmentation by highlighting the intra-annual variation characteristic of active crop fields~\cite{estes2022high,debats2016generalized}. This contrast can help models rule out potential false positives such as fallow fields or forest stands. 

For each country, we collected images from two date ranges (Table~\ref{tab:dataset-details}). Growing seasons vary greatly globally (and even within countries). To choose the date ranges, we looked at each country's crop calendar which specifies the planting, mid-season, harvesting, and off-season months for the main crops~\cite{usda_ipad_crop_calendar}. We inspected the available satellite images in the sample area(s) in two date ranges spanning the planting/mid-season and harvesting/off-season months. If a country had multiple growing seasons (e.g., winter and summer crops), we visualized both seasons and chose the one appearing most active. We then iteratively adjusted the date ranges to account for good contrast between images and cloud cover. After adjustment, the date ranges do not necessarily match the growing season stages, so we call them Window A and B. 

\subsubsection{Cloud filtering}
For each chip ROI, we searched for Sentinel-2 scenes with $<\! 90\%$ scene level cloud cover in the two date ranges. For the resulting scenes, we cropped each scene to the sample chip ROI and computed the cloud percentage in the patch using the Sentinel-2 scene classification layer (SCL) ``Cloud medium probability'' and ``Cloud high probability'' classes. We selected the chip with the lowest cloud percentage. If there were no chips with cloud percentage $<\! 10\%$, we discarded the chip. Table~\ref{tab:dataset-details} gives the resulting number of Sentinel-2 chips created for each country.

\subsubsection{Resizing and normalization}
We resized each chip to $256\times256$ pixels. Each chip is stored as a GeoTIFF with EPSG:4326. 
We normalize images during training by dividing each channel by $3000$ (an approximate mean value).

\subsection{Dataset splits} \label{subsec:data_splits}
\ftw defines training, validation, and test sets for each country to facilitate evaluation of test metrics at the country scale. 
Many sample chips are spatially adjacent since they were tiled from large grids. %
Spatial autocorrelation between adjacent chips may cause leakage between data subsets if chips are randomly split into subsets~\cite{rolf2023evaluation}. %
To reduce the impact of spatial autocorrelation, we implemented a blocked random splitting strategy. We grouped chips into $3\times3$ blocks and randomly assigned 80\% to training, 10\% validation, and 10\% test.
To ensure $3\times3$ blocks were large enough, we performed an experiment to quantify the sensitivity of test performance to the block size and distance from each test patch to the nearest training patch (see supplement). 

\section{Dataset Analysis and Related Work}
\label{sec:stats}

\begin{table*}[]
\caption{Key attributes of Fields of The World and previous field boundary segmentation datasets.}
\label{tab:relatedwork}
\centering
\resizebox{\textwidth}{!}{%
\begin{threeparttable}
\begin{tabular}{@{}cccccccc@{}}
\toprule
\textbf{Dataset} & \textbf{\# countries} & \textbf{Sensor(s)} & \textbf{\begin{tabular}[c]{@{}c@{}}Input dim\\ ([$H, W, C, T$])\end{tabular}} & \textbf{\begin{tabular}[c]{@{}c@{}}\# field polygons\\ (million)\end{tabular}} & \textbf{\# total samples} & \textbf{\begin{tabular}[c]{@{}c@{}}Sample dim.\\ (m$\times$m)\end{tabular}} & \textbf{\begin{tabular}[c]{@{}c@{}}Total area\\ (km$^2$)\end{tabular}} \\ \midrule
\ftwlong (\ftw) & \totalcountries & Sentinel-2 & $[256,256,4,2]$ & 1.63 & 70,484 & 1,536 & 166,293 \\ \midrule
AI4Boundaries \cite{d2023ai4boundaries} & 7 & \begin{tabular}[c]{@{}c@{}}Sentinel-2,\\ Aerial\end{tabular} & \begin{tabular}[c]{@{}c@{}}$[256,256,4,12]$\\ $[512,512,3,1]$\end{tabular} & 1.07\tnote{3} & \begin{tabular}[c]{@{}c@{}}7,831\\ 7,598\end{tabular} & \begin{tabular}[c]{@{}c@{}}2,560\\ 512\end{tabular} & \begin{tabular}[c]{@{}c@{}}51,321\\ 1,992\end{tabular} \\ \midrule
AI4SmallFarms \cite{persello2023ai4smallfarms} & 2 & Sentinel-2 & $[256,256,4,1]$ & 0.44 & 62 & 5,000 & 1,550 \\ \midrule
PASTIS \cite{garnot2021panoptic} & 1 & Sentinel-2 & $[128,128,10,N]\tnote{1}$ & 0.12 & 2,433 & 1,280 & 3,986 \\ \midrule
PASTIS-R \cite{garnot2022multi} & 1 & \begin{tabular}[c]{@{}c@{}}Sentinel-2,\\ Sentinel-1\end{tabular} & \begin{tabular}[c]{@{}c@{}}$[128,128,10,N]\tnote{2}$\\ $[128,128,3,70]$\end{tabular} & 0.12 & 2,433 & 1,280 & 3,986 \\ \bottomrule
\end{tabular}%
\begin{tablenotes}
\item[1] Varying observations (38-61) taken between September 2018 and November 2019.
\item[2] All available observations for the 2019 season.
\item[3] \citet{d2023ai4boundaries} reports 2.5M parcels contained in 7,831 4-km samples, however the number of polygons included in dataset sample masks is smaller. 
\end{tablenotes}
\end{threeparttable}
}
\end{table*}

\begin{figure}[t!]
    \centering
    \includegraphics[width=0.9\linewidth]{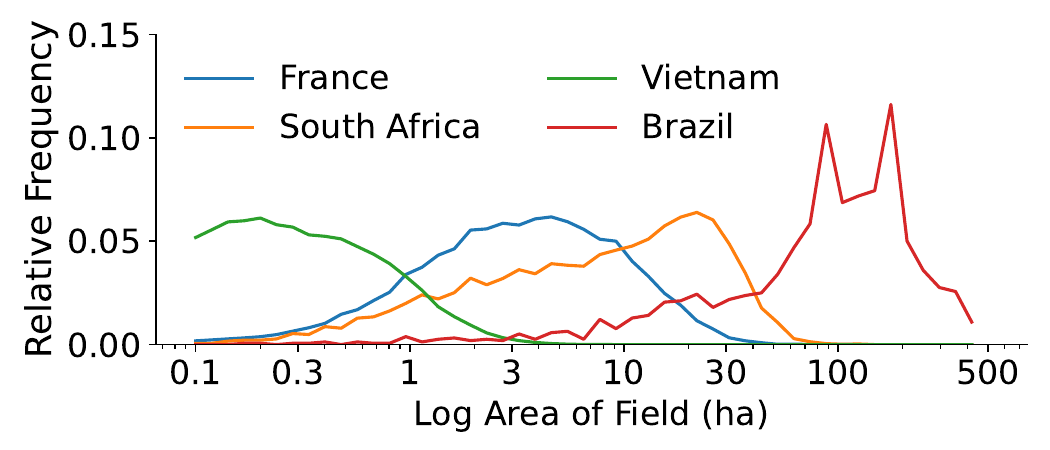}
    \caption{Field area distribution across four countries.}
    \label{fig:area}
\end{figure}

The dramatic differences in field morphology across the globe motivated the construction of \ftw (Figure~\ref{fig:teaser}). \ftw has significant advantages over previous field instance segmentation datasets in terms of (i) geographic representation and extent, (ii) annotation volume, and (iii) annotation/scene complexity. The most relevant datasets for comparison are AI4Boundaries~\cite{d2023ai4boundaries}, AI4SmallFarms~\cite{persello2023ai4smallfarms}, PASTIS~\cite{garnot2021panoptic}, and PASTIS-R~\cite{garnot2022multi}. We only include datasets that explicitly label individual field instances, excluding semantic segmentation labels since field instances enable a broader range of applications. We also exclude datasets providing field polygons but no imagery, such as the field boundary dataset created by the French Land Parcel Identification System~\cite{rpg}. The lack of standardized satellite imagery for polygon-only datasets hampers the creation and comparison of automated algorithms for field boundary delineation and related tasks.

Table~\ref{tab:relatedwork} compares the key attributes for all datasets (see geographic distributions in the supplement). \ftw has a significantly broader geographic distribution than all previous datasets, including fields from \totalcountries countries spanning four continents (Europe, Asia, Africa, and South America). Previous datasets include at most 7 countries and are mostly concentrated in Europe, except AI4SmallFarms which only has images from Cambodia and Vietnam. \ftw is the largest dataset in terms of number of samples, total area, and total annotations. \ftw has an order of magnitude more samples and area covered than the next-largest dataset (AI4Boundaries~\cite{d2023ai4boundaries}), with \patchtotal sample chips spanning 166,293 \unit{\square\km}. It also has the highest annotation volume with 1.63M field polygons, compared to the next-largest volume of 1.07M  in AI4Boundaries. %

The \ftw dataset captures greater morphological diversity of agricultural field instances than any other dataset. Figure~\ref{fig:teaser} shows example field boundaries from four countries. The diverse shapes and sizes reflect the unique topographical, environmental, and historical factors that influenced the development of field boundaries in each region. Figure~\ref{fig:area} shows the dramatic difference between the field areas in different countries. There is little overlap between the distributions of Vietnam (small fields) and Brazil (large fields). Figure~\ref{fig:area-others} shows the distribution field areas for AI4SmallFarms, AI4Boundaries, and \ftw (note that we did not include PASTIS/PASTIS-R since both provide field boundaries in raster, not vector, format so we could not compute morphological statistics). AI4SmallFarms consists mostly of small-area fields,  AI4Boundaries consists mostly of medium-area fields, and \ftw has a broader distribution of field areas.
There is also significant diversity in field shape complexity (see visualizations in supplement). For example, Estonia and Slovakia have complex field shapes, with 111.9 and 95.4 polygon vertices on average, respectively. In contrast, Kenya and Rwanda have simpler field shapes, with 4.5 and 5.6 vertices on average, respectively.

\ftw provides a more complete representation of the diversity and complexity of agricultural landscapes across the globe than previous datasets. We hope that \ftw will lead to more broadly applicable models for field boundary segmentation by providing a large and diverse training dataset and enabling region-specific evaluation and error analysis.

\begin{table*}[ht!]
\centering
\caption{Performance metrics for different target mask formats in Slovenia (SVN), France (FRA), and South Africa (ZAF). We compared 2-class field extent and 3-class masks with or without ignoring background (bg) pixels for presence-only samples.}
\resizebox{\linewidth}{!}{%
\begin{tabular}{@{}lccccccccccccccc@{}}
\toprule
\multicolumn{1}{c}{\multirow{2}{*}{\textbf{Mask type}}} & \multicolumn{3}{c}{Pixel IoU} & \multicolumn{3}{c}{Pixel precision} & \multicolumn{3}{c}{Pixel recall} & \multicolumn{3}{c}{Object precision} & \multicolumn{3}{c}{Object recall} \\ \cmidrule(l){2-4}\cmidrule(l){5-7} \cmidrule(l){8-10}\cmidrule(l){11-13} \cmidrule(l){14-16} 
\multicolumn{1}{c}{} & SVN & FRA & ZAF & SVN & FRA & ZAF & SVN & FRA & ZAF & SVN & FRA & ZAF & SVN & FRA & ZAF \\ \midrule
2-class & 0.66 & \textbf{0.83} & \textbf{0.83} & 0.84 & 0.87 & 0.87 & 0.76 & 0.95 & 0.94 & 0.52 & 0.64 & 0.54 & 0.06 & 0.29 & 0.24 \\
2-class (ignore presence-only bg) & \textbf{0.69} & 0.82 & 0.81 & 0.78 & 0.86 & 0.84 & \textbf{0.86} & \textbf{0.95} & \textbf{0.96} & 0.30 & 0.38 & 0.44 & 0.08 & 0.15 & 0.19 \\
\midrule
3-class & 0.67 & \textbf{0.83} & \textbf{0.83} & 0.87 & 0.88 & 0.88 & 0.75 & 0.94 & 0.94 & \textbf{0.60} & \textbf{0.71} & \textbf{0.63} & 0.07 & 0.45 & 0.35 \\
3-class (ignore presence-only bg) & 0.59 & 0.79 & 0.79 & \textbf{0.90} & \textbf{0.89} & \textbf{0.89} & 0.63 & 0.88 & 0.88 & 0.33 & 0.55 & 0.51 & \textbf{0.20} & \textbf{0.58} & \textbf{0.55} \\ \bottomrule
\end{tabular}%
}
\label{tab:exp1-masks}
\end{table*}
\begin{table*}[t]
\caption{Ablation results for multispectral (RGB-NIR vs.~RGB only) and multi-temporal (Window A and Window B) input channels in Slovenia (SVN), France (FRA), and South Africa (ZAF).}
\resizebox{\textwidth}{!}{%
\begin{tabular}{@{}lrrrrrrrrrrrrrrr@{}}
\toprule
\multicolumn{1}{c}{\multirow{2}{*}{\textbf{Channels}}} & \multicolumn{3}{c}{Pixel IoU} & \multicolumn{3}{c}{Pixel precision} & \multicolumn{3}{c}{Pixel recall} & \multicolumn{3}{c}{Object precision} & \multicolumn{3}{c}{Object recall} \\ \cmidrule(l){2-4}\cmidrule(l){5-7} \cmidrule(l){8-10}\cmidrule(l){11-13} \cmidrule(l){14-16}  
\multicolumn{1}{c}{} & \multicolumn{1}{l}{SVN} & \multicolumn{1}{l}{FRA} & \multicolumn{1}{l}{ZAF} & \multicolumn{1}{l}{SVN} & \multicolumn{1}{l}{FRA} & \multicolumn{1}{l}{ZAF} & \multicolumn{1}{l}{SVN} & \multicolumn{1}{l}{FRA} & \multicolumn{1}{l}{ZAF} & \multicolumn{1}{l}{SVN} & \multicolumn{1}{l}{FRA} & \multicolumn{1}{l}{ZAF} & \multicolumn{1}{l}{SVN} & \multicolumn{1}{l}{FRA} & \multicolumn{1}{l}{ZAF} \\ 
\midrule
Stacked Windows A and B & \textbf{0.58} & \textbf{0.79} & \textbf{0.80} & \textbf{0.91} & \textbf{0.89} & \textbf{0.90} & 0.61 & \textbf{0.87} & 0.87 & \textbf{0.30} & \textbf{0.54} & \textbf{0.55} & \textbf{0.18} & \textbf{0.58} & \textbf{0.54} \\
Stacked Windows A and B (RGB only) & \textbf{0.58} & 0.78 & 0.79 & 0.90 & \textbf{0.89} & 0.89 & \textbf{0.62} & 0.86 & \textbf{0.88} & 0.27 & 0.51 & 0.53 & \textbf{0.18} & 0.56 & \textbf{0.54} \\
Mean of Windows A and B & 0.54 & 0.77 & 0.78 & 0.88 & \textbf{0.89} & 0.88 & 0.59 & 0.86 & \textbf{0.88} & 0.27 & 0.50 & 0.49 & 0.16 & 0.55 & 0.53 \\
Window A only & 0.55 & 0.77 & 0.78 & 0.88 & 0.88 & 0.88 & 0.59 & 0.86 & 0.87 & 0.27 & 0.47 & 0.47 & 0.17 & 0.54 & 0.52 \\
Window B only & 0.52 & 0.78 & 0.79 & 0.87 & \textbf{0.89} & 0.89 & 0.57 & 0.86 & 0.87 & 0.24 & 0.49 & 0.52 & 0.15 & 0.54 & 0.53 \\ \bottomrule
\end{tabular}%
}
\label{tab:exp1-channels}
\end{table*}

\begin{table*}[ht]
\centering
\caption{Transfer learning results for models pre-trained on France (FRA) or Netherlands (NLD), AI4Boundaries countries (Austria/AUT, Spain/ESP, FRA, Luxembourg/LUX, NLD, Slovenia/SVN, and Sweden/SWE), or \ftw minus the target region. Models are fine-tuned and tested on the target region. We report recall metrics only for India since it has presence--only labels. Each cell gives two results: no fine-tuning / after fine-tuning using the target training set.}
\resizebox{\textwidth}{!}{%
\begin{tabular}{@{}lclccccc@{}}
\toprule
\multicolumn{1}{c}{\textbf{Analogous work}} & \textbf{Fine--tune/test} & \multicolumn{1}{c}{\textbf{Pre--train}} & \textbf{Pixel IoU} & \textbf{Pixel precision} & \textbf{Pixel recall} & \textbf{Object precision} & \textbf{Object recall} \\ \midrule
\multirow{3}{*}{\citet{wang2022unlocking}} & \multirow{3}{*}{India} & FRA & 0.51 / 0.50 & - & 0.60 / 0.55 & - & 0.03 / 0.13 \\
 &  & AUT, ESP, FRA, LUX, NLD, SVN, SWE & 0.16 / 0.54 & - &  0.16 / 0.59 & - & 0.05 / 0.16 \\
 &  & \ftw $-$ \{India\} & \textbf{0.57} / \textbf{0.55} & - & \textbf{0.63} / \textbf{0.60} & - & \textbf{0.14} / \textbf{0.19} \\ \midrule
\multirow{3}{*}{\citet{persello2023ai4smallfarms}} & \multirow{3}{*}{\begin{tabular}[c]{@{}c@{}}Cambodia\\ Vietnam\end{tabular}} & NLD & 0.04 / 0.48 & 0.77 / 0.90 & 0.04 / 0.51 & 0.02 / 0.22 & 0.00 / 0.18 \\
 &  & AUT, ESP, FRA, LUX, NLD, SVN, SWE & 0.16 / 0.51 & 0.90 / 0.90 & 0.16 / 0.54 & 0.12 / 0.27 & 0.04 / 0.22 \\
 &  & \ftw $-$ \{Cambodia, Vietnam\} & \textbf{0.43} / \textbf{0.55} & \textbf{0.92} / \textbf{0.91} & \textbf{0.44} / \textbf{0.58} & \textbf{0.21} / \textbf{0.29} & \textbf{0.16} / \textbf{0.24} \\ \bottomrule 
\end{tabular}%
}
\label{tab:exp2-pretraining}
\end{table*}

\section{Baseline Experiments}
\label{sec:baselines}

\paragraph{Setup and metrics}
We follow the common approach to field instance segmentation of segmentation 
then polygonization of predicted raster masks~\cite{persello2023ai4smallfarms}. Unless specified otherwise, we used a U-net with EfficientNet-b3 backbone with inputs consisting of concatenated 4-channel RGB-NIR images from Window A and B. We found that this architecture performed well compared to other architectures and backbones (see \textbf{Architectures} paragraph in this section and Table~\ref{tab:exp1-architecture} in supplement). We also found that concatenating both temporal windows and using four spectral bands performed best compared to other configurations (see \textbf{Multi-temporal and multispectral channels} paragraph and Table~\ref{tab:exp1-channels}). We initialized the RGB channels using ImageNet and NIR channels using random weights. We optimized cross-entropy loss with class weights inversely proportional to each class's frequency in the training set. We trained all models for 100 epochs. We used a fixed random seed for all experiments (randomly chosen). We did not perform hyperparameter tuning. Experiments required 4 A100 and 8 V100 GPUs for approximately one week.

We used semantic (pixel-level) and instance (object-level) segmentation metrics: pixel-level intersection over union (IoU), precision, and recall, and object-level precision and recall (functions in the \ftw code repository). We converted segmentation masks to polygons using \texttt{rasterio} then computed object-level metrics with IoU threshold of $0.5$. %

\paragraph{Modeling configuration}
\ftwlong includes multiple target mask formats and satellite images from two dates and four spectral channels, giving users many modeling choices. In semantic segmentation followed by polygonization, there are other choices such as which architecture to choose. We performed several experiments to assess the impact of these choices on model performance. In each experiment, we evaluate performance using the test set for each country. We report results for countries with presence/absence labels chosen to span a range of field sizes: Slovenia - SVN (average field size $0.64$ ha), France - FRA (average $5.7$ ha), and South Africa - ZAF (average $13.8$ ha). We provide results for all countries in the supplementary material. %
An extensive search of modeling configurations is beyond the scope of this work, but we hope future studies will explore a greater range of options using the \ftw dataset.

\begin{figure}[t!]
    \centering
    \includegraphics[width=\columnwidth]{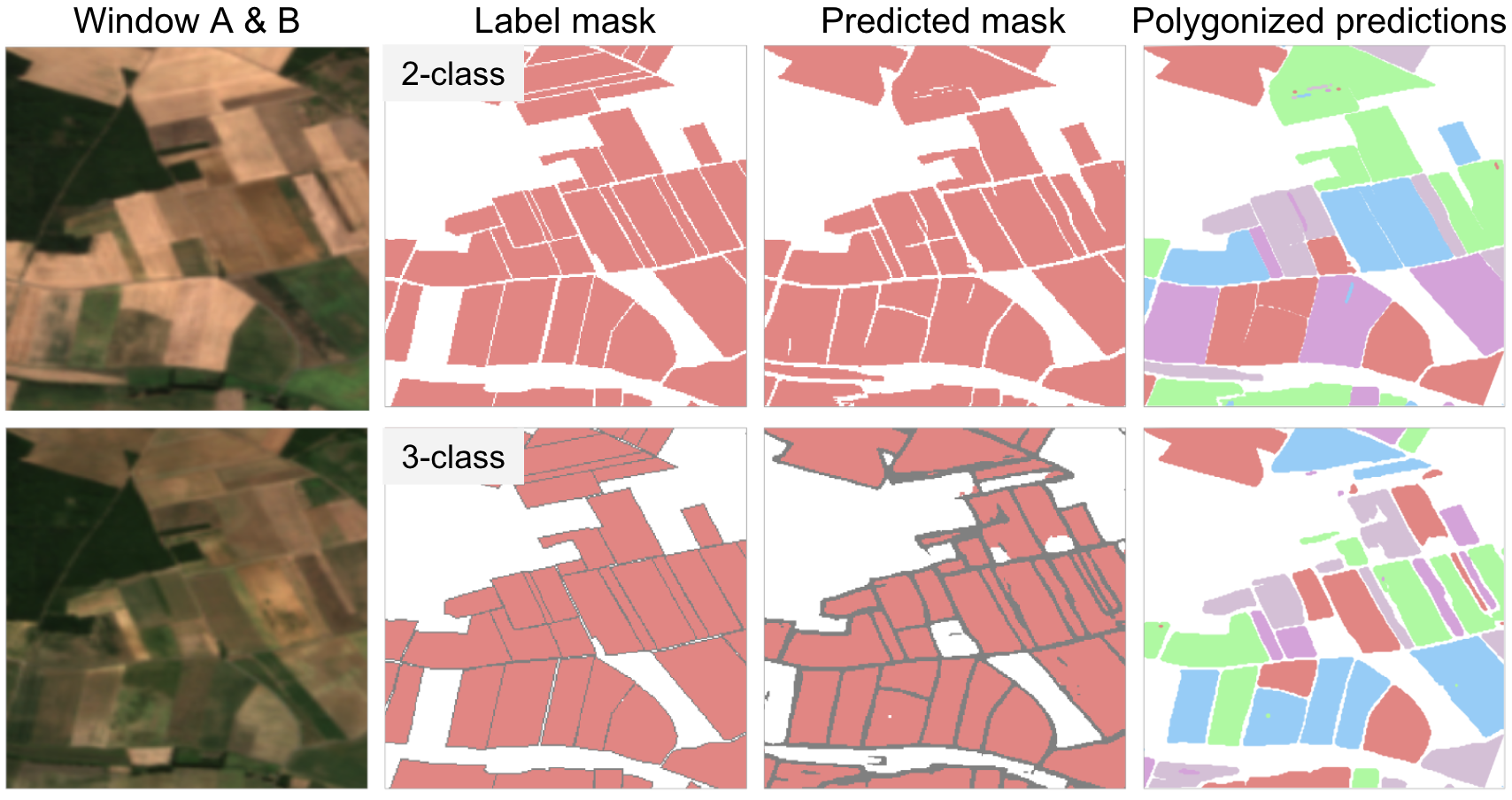}
    \caption{Example France predictions for 2-class and 3-class models in rows 2 and 4 of Table~\ref{tab:exp1-masks}.}
    \label{fig:visual-results}
\end{figure}

\paragraph{Target mask}
We evaluated the two types of target masks provided in \ftw: 2-class (field interior vs.\ background) and 3-class (field interior, boundary, and background). In presence-only countries, pixels with an unknown class are labeled as background. We evaluated two scenarios: 1) when computing the loss, ignore all pixels labeled background for presence-only countries, and 2) compute the loss for all pixels treating unknown labels as background. 
Before computing test metrics, we converted outputs from all models to binary field extent masks to ensure a common evaluation basis.

Rows 1 and 3 of Table~\ref{tab:exp1-masks} show that almost all metrics are higher with 3-class masks than binary masks. 
Rows 2 and 4 compare when unknown-label pixels are counted as background or ignored for presence-only samples when training with 3-class masks. Although object precision is lower when ignoring unknown pixels, object recall is significantly higher, especially for Slovenia. Figure \ref{fig:visual-results} shows that the segmentation masks predicted by both models are good, but 3-class masks improve the delineation of contiguous fields and thus object recall. From these results, we concluded that training with 3-class masks and ignoring presence-only background is most likely to give good performance across all regions and used this setup for subsequent experiments.

\paragraph{Multi-temporal and multispectral channels}
We did an ablation experiment to evaluate the benefit of the two contrasting image dates (Window A and B) in \ftw. We also evaluated a mean of both windows. Finally, we evaluated with/without the NIR channel.
Table~\ref{tab:exp1-channels} shows the best performance comes from both time windows and all spectral channels. Overall, removing one of the time windows causes a greater drop in metrics than removing the NIR channel. This is consistent with previous work that showed performance improvements were greater when adding more timesteps compared to more spectral channels~\cite{debats2016generalized}.

\paragraph{Architecture}
We evaluated U-net~\cite{ronneberger2015u} and DeepLabv3+~\cite{chen2018encoder} models with 5 different backbones: ResNet-18, ResNet-50, ResNeXt-50, EfficientNet-b3, and EfficientNet-b4. Overall performance is similar across different architectures and backbones, though U-net models tend to outperform DeepLabv3+ models (results in supplement). 

\paragraph{Transfer learning}
Some countries have many labels while others have few (Table~\ref{tab:dataset-details}). Prior work showed that performance on a data-scarce region could be improved by pre-training models on a country with a large labeled dataset and then fine-tuning on the target region.~\citet{wang2022unlocking} fine-tuned a model for India after pre-training on France.~\citet{persello2023ai4smallfarms} fine-tuned for Cambodia and Vietnam after pre-training on the Netherlands. 

Direct comparisons to these works are not possible due to differences in data formats. Instead, we performed three analogous experiments to evaluate the improvement of pre-training on \ftw compared to smaller, more geographically limited datasets as in prior work: 1) pre-training on one data-rich country (France or Netherlands), 2) pre-training on the countries included in AI4Boundaries (to emulate pre-training on AI4Boundaries), and 3) pre-training on \ftw with the target country held-out. We then fine-tuned each model for 200 epochs using the target country (India or Cambodia+Vietnam) training set and evaluated on its test set.  

Table~\ref{tab:exp2-pretraining} shows that models pre-trained with \ftw outperform models trained on more geographically limited subsets in both target regions. The performance of \ftw models without any fine-tuning is especially impressive. \ftw models with no fine-tuning perform similarly or better than fully fine-tuned versions of compared models.

\begin{figure}[t]
    \centering
    \includegraphics[width=0.95\columnwidth]{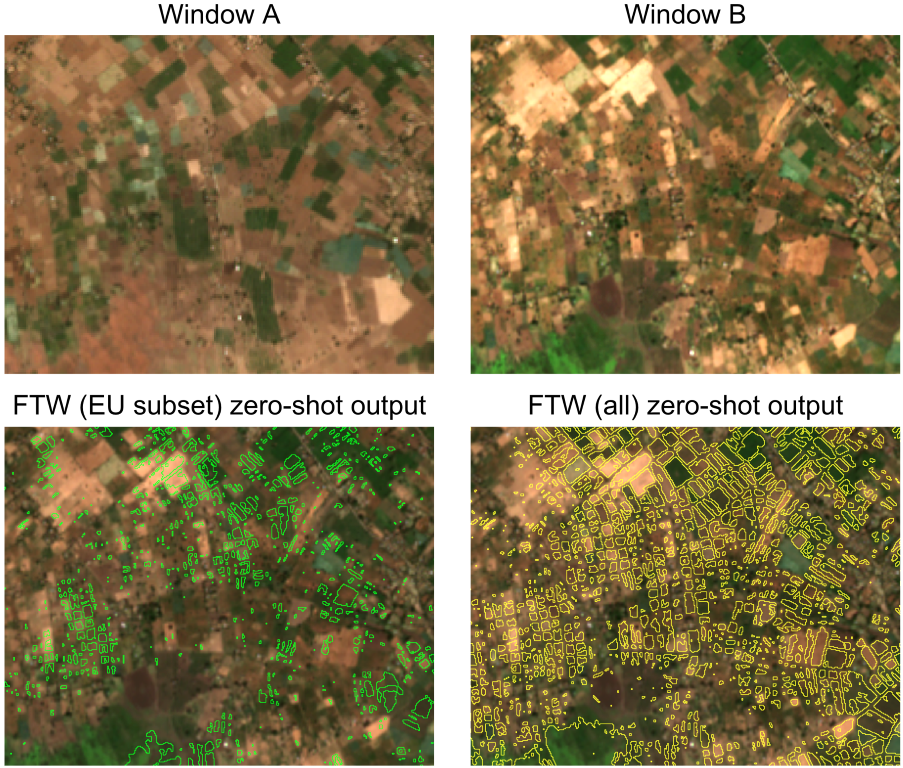}
    \caption{Zero-shot predictions with no post-processing for a $20$ sq km region in Ethiopia ($8^\circ 05'$N,\; $38^\circ 51'$E). A \ftw pre-trained model achieves (qualitatively) good performance, even though Ethiopia is not in the \ftw training set and is a challenging region to delineate fields.}
    \label{fig:inference}
\end{figure}

\paragraph{Deployment readiness}
Motivated by the impressive performance of \ftw pre-trained models without fine-tuning in Table~\ref{tab:exp2-pretraining}, we used a \ftw pre-trained model to predict field boundaries in Ethiopia, a challenging region not in \ftw (Figure \ref{fig:inference}). The results show good qualitative performance that could be improved with local fine-tuning and post-processing. This shows the high potential of \ftw to be immediately used in practice with little adaptation effort.

\section{Discussion and Conclusion}
ML research on automatic extraction of agricultural field boundaries from remotely sensed imagery is limited by a lack of ML-ready datasets to train and evaluate models on the global diversity of crop fields. These datasets are urgently needed in many applications for agriculture, climate change, and development. 
We designed \ftwlong to improve ML model performance for field boundary segmentation in diverse global agricultural landscapes and enable granular country-scale evaluation for more countries than any prior dataset. Our experiments established a performance baseline for the new \ftw benchmark and showed that \ftw-trained models perform better than more geographically limited datasets analogous to existing benchmarks. 

Future work could build on these baselines by testing more model architectures, including instance segmentation architectures (e.g., Mask-RCNN~\cite{he2017mask} or SAM~\cite{kirillov2023segment}) and geospatial foundation models (e.g., SatMAE~\cite{cong2022satmae} or Presto~\cite{tseng2023lightweight}). Future work could also explore other methods of constructing target masks, motivated by our result that training with 3-class masks performed better than 2-class masks. 

We provide complete metadata for sample grids, sample chips, and field boundary polygons to enable future extensions of \ftwlong. For example, future work could add more spectral channels or sensors, timesteps, or sample locations. \ftw can be extended as field polygons become available for more countries. We hope the community will build on \ftw as research on this important task grows. 

\paragraph{Benchmarking on \ftw} We hope this study will inspire researchers to develop new methods for field boundary segmentation and measure improvement using the \ftw benchmark. We suggest benchmarking performance on the per-country test sets and reporting individual country results, the mean across all countries, or the minimum across countries (worst-case performance). Supplement Table \ref{tab:benchmarking-example} reports these metrics for the best model evaluated in this paper. 

\section*{Ethics statement}
Researchers, practitioners, and other users of \ftwlong must be aware of important ethical considerations raised by the digitization of field boundaries and other information from publicly accessible satellite data. Digitized field boundary data could inadvertently expose the practices and characteristics of individual land parcels, which could infringe on the privacy of local landowners who may be unaware of this digitization or its implications. There are also risks that private or public entities may use digitized field boundary data in a way that marginalizes vulnerable individuals such as smallholder farmers.~\citet{rolfposition} summarized distinct ethical concerns of machine learning applied to satellite data. In line with the recommendations of~\citet{rolfposition}, we suggest that users of \ftw work with local organizations and communities to build and release responsible field boundary datasets and ensure their project goals and practices align with local needs and regulations.

\section*{Acknowledgments}
This project was supported by funding from the Taylor Geospatial Engine and a NASA Supplemental Open Source Software Award.

\bibliography{refs}

\clearpage
\newpage

\appendix

\section{Supplementary Information}

\subsection{Annotation filtering}

\subsubsection{Additional details on semantic filtering}
\begin{figure}[ht]
    \centering
    \begin{subfigure}[t]{0.9\linewidth}
        \centering
        \includegraphics[width=\linewidth]{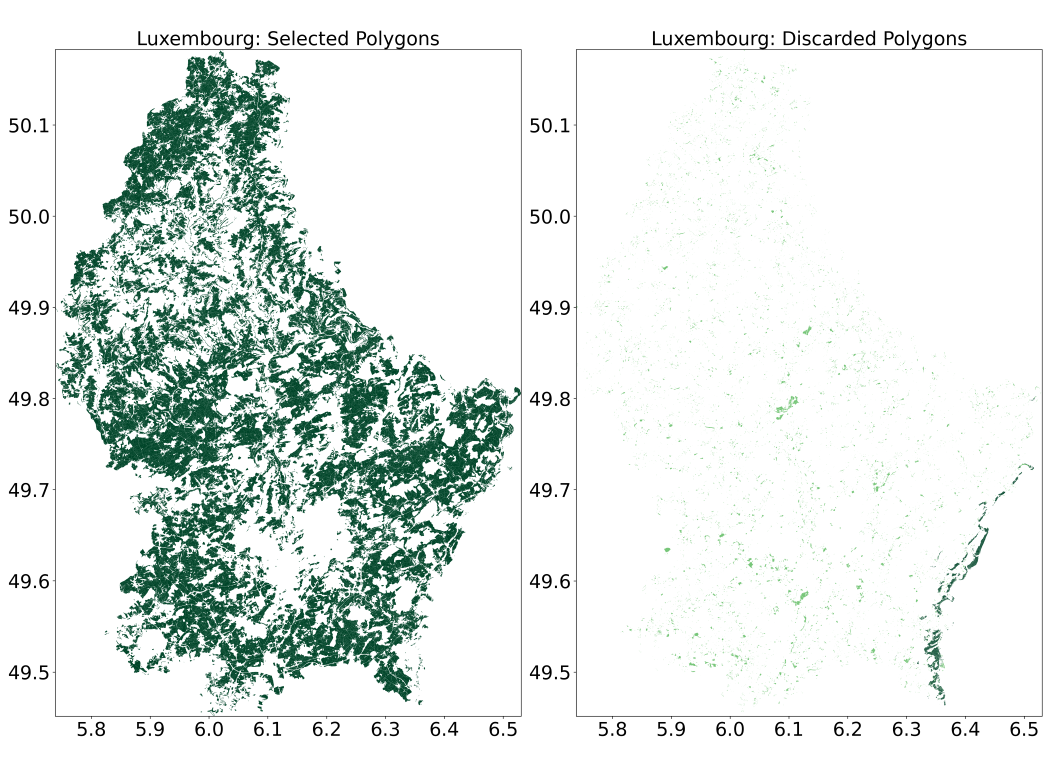}
        \caption{Selected and Discarded Crop Polygons in Luxembourg}
        \label{fig:luxembourg-filter-example}
    \end{subfigure}
        \vspace{1em} %
    \begin{subfigure}[t]{\linewidth}
        \centering
        \includegraphics[width=\linewidth]{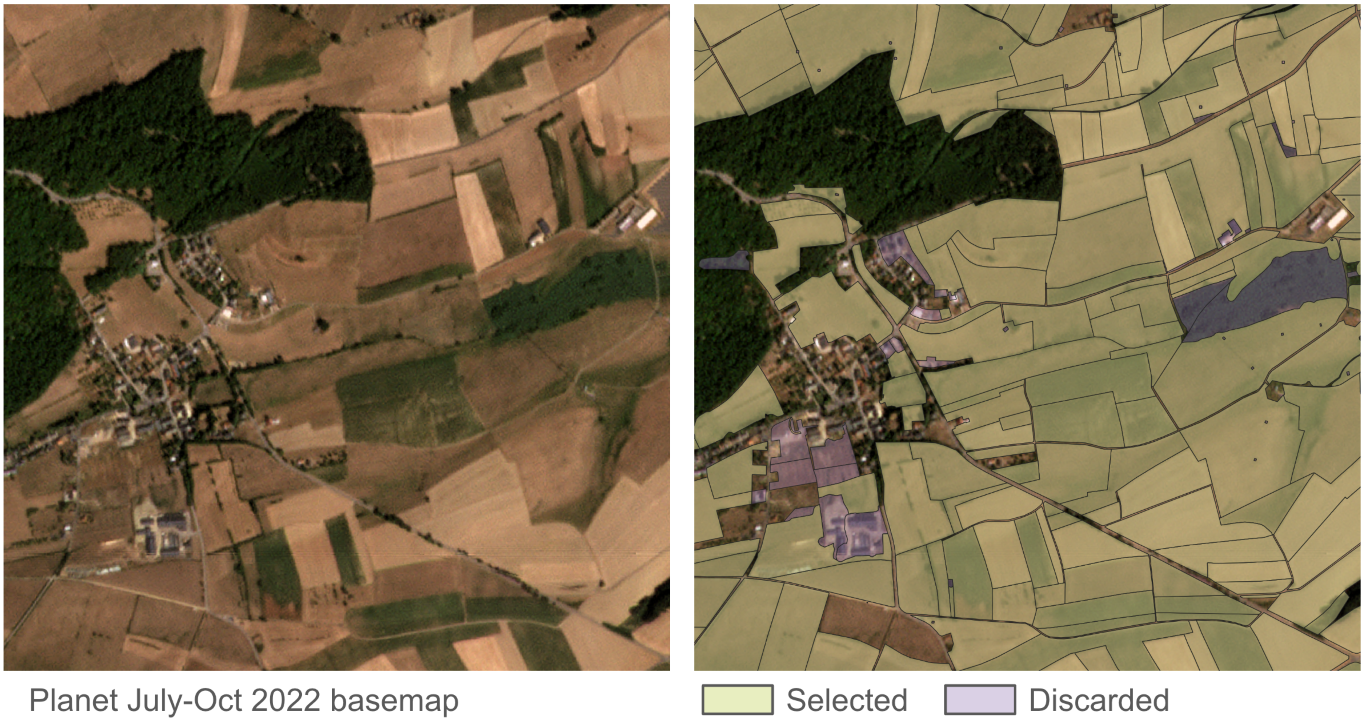}
        \caption{Zoomed-in visualization of selected (green) and discarded (purple) polygons in Luxembourg.}
        \label{fig:filter-example}
    \end{subfigure}

    \caption{Visualization of polygons that were included (selected) and filtered out (discarded) in the Luxembourg dataset.}
    \label{fig:combined-filter-example}
\end{figure}

As described in the \textit{Semantic filtering} section, we excluded all classes that were not annual (temporary) crops if they were included in a field boundary dataset. In \texttt{ftw-semantic-filters.csv} (found at \url{https://github.com/fieldsoftheworld/ftw-datasets-list}), we list and justify the classes included and excluded in each dataset. We also give the exact dates used to filter each country's satellite images for Window A and Window B. In Figure \ref{fig:combined-filter-example}, we visualize the selected and discarded (filtered-out) fields in Luxembourg.

\begin{figure}[ht]
    \centering
    \includegraphics[width=\linewidth]{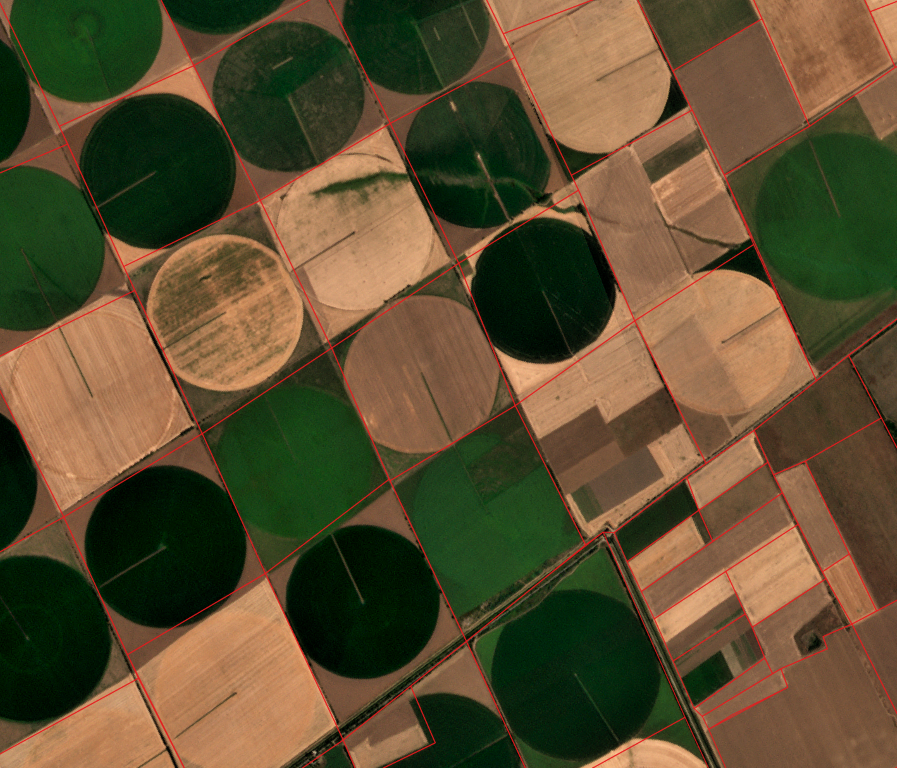}
    \caption{Example field boundaries in the GloCAB dataset~\cite{hall2024glocab}. Some boundaries did not align with the boundary apparent in satellite images, especially for center-pivot fields in Ukraine.}
    \label{fig:glocab}
\end{figure}

\subsubsection{Field boundary datasets not included in \ftw}

We conducted a comprehensive search for field polygons from government databases, published literature, and other websites to use as annotations in \ftw.
We looked for datasets with diverse geographic coverage, high-quality and trustworthy polygon annotations, and licenses that permit reuse. We included all datasets meeting these criteria in \ftw. We considered author-reported quality assessment in each dataset's documentation, previous use of the dataset in ML analyses, and visual inspection (e.g., closed polygons, polygons consistent with satellite images from reported dates, etc). 

There were a few datasets that did not meet our criteria and thus we decided not to include in \ftw: 
\begin{itemize}
    \item Zambia: The same source data provider of our Kenya dataset, ECAAS, also published a dataset in Zambia. The polygons in this dataset were extremely sparse and did not appear to align with satellite imagery from the same year.The dataset can be obtained from \url{https://drive.google.com/drive/folders/1nEhHxWzsZxqozO2LZa-uUl6DoKNVYZVZ} and metadata from \url{https://ecass-project-documentation.readthedocs.io/en/latest/modules/data_access.html}.
    \item Romania: This documentation of this dataset did not specify the year the field boundaries were valid for. We decided not to use the dataset because we did not know what year of satellite imagery it should be paired with. The dataset can be found at \url{https://github.com/maja601/EuroCrops/wiki/Romania}.
    \item Kenya: \citet{kehs2021village} published a crop type dataset with field boundary polygons in Kenya. However, the paper describes limited quality assessment and our visual inspection showed some fields did not align well with with contemporaneous satellite imagery.
    \item Brazil: The Cadastro Ambiental Rural (CAR) (\url{https://dados.agricultura.gov.br/it/dataset/cadastro-ambiental-rural}) provides geo-referenced data for land parcels including agriculture~\cite{jung2017brazil}. However, we were not able to determine the appropriate attributes and attribute values to determine how to filter the parcels for temporary crops. We will try to obtain this information in future work to include in a later version of \ftw. 
    \item GloCAB (Brazil, Ukraine, USA, Canada, and Russia): \citet{hall2024glocab} published the GloCAB of 190,832 manually-digitized field boundaries in 22 regions of various sizes spanning 5 countries: Brazil, Ukraine, United States of America, Canada, and Russia. While this dataset seems promising for inclusion in \ftw, visual inspection revealed many boundaries that did not align with the apparent field extent from the satellite imagery, particularly around center-pivot irrigated fields in Ukraine (see Figure~\ref{fig:glocab}). We will continue to investigate using this dataset in future work and hope to include a filtered version of it in future \ftw versions.
    \item USA (California): The Kern County Department Of Agriculture And Measurement publishes crop field boundaries annually since 1997. We were not able to include this dataset in \ftw because their website does not specify a license for the data that allows for reuse.
\end{itemize}

\subsection{Effect of random spatial splits}
\label{sec:random_splits}

\begin{figure}[ht]
    \centering
    \includegraphics[width=0.7\linewidth]{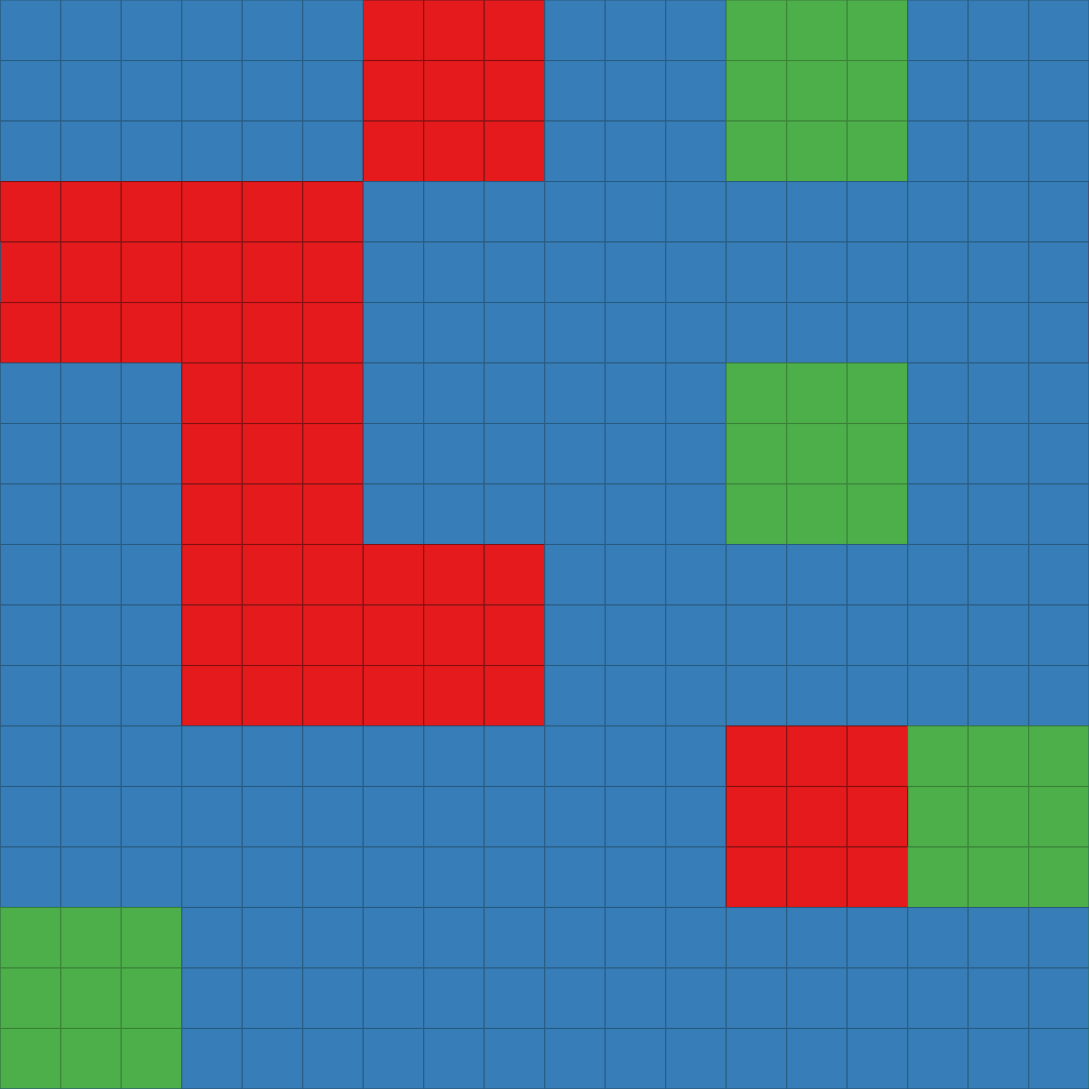}
    \caption{Example of block splits in France where red patches are in the test set, blue patches are in the train set, and green patches are in the validation set.}
    \label{fig:france_blocks}
\end{figure}

As described in the \textit{Dataset splits} section, we perform a blocked random splitting strategy to partition $3\times3$ groups of patches into training, validation, and test splits for each country in the dataset. Figure~\ref{fig:france_blocks} shows an example of this splitting strategy for a section of the France dataset. As such, patches in the test sets are adjacent to patches in the train sets, which may allow for leakage between train and test due to spatial autocorrelation in imagery in labels. 

We tested for this effect by grouping test patches by the number of training patches they are adjacent to, then computing model performance for each group (using the entire \ftw dataset). If autocorrelation was causing data leakage, then we would expect to observe higher model performance among test patches that are adjacent to training patches compared to test patches that are isolated (e.g., in the middle of the 3x3 blocks). We compared the distribution of Pixel IoU per patch using the 2-class (ignore presence-only background) model between the group with no adjacent training patches to those with some adjacent training patches per country with an independent sample t-test. We did not find a statistically significant difference in performance for any country and concluded that spatial autocorrelation is not influencing test set results.

\subsection{Dataset characteristics}

In this section, we provide additional dataset visualizations to show the diversity in field morphology between countries and within the \ftwlong dataset. 

\begin{figure}[ht]
    \centering
    \includegraphics[width=0.9\linewidth]{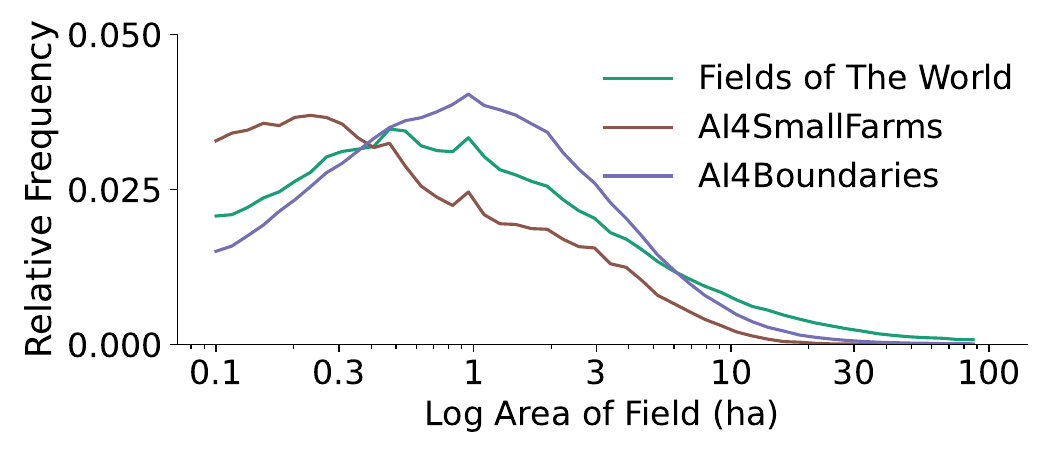}
    \caption{The distribution of (log) field area in \ftw and two previous benchmark datasets.}
    \label{fig:area-others}
\end{figure}

Figure~\ref{fig:area-others} shows the distribution of (log) field area across \ftw and two previous benchmark datasets.

\begin{figure}[ht]
    \centering
    \includegraphics[width=\columnwidth]{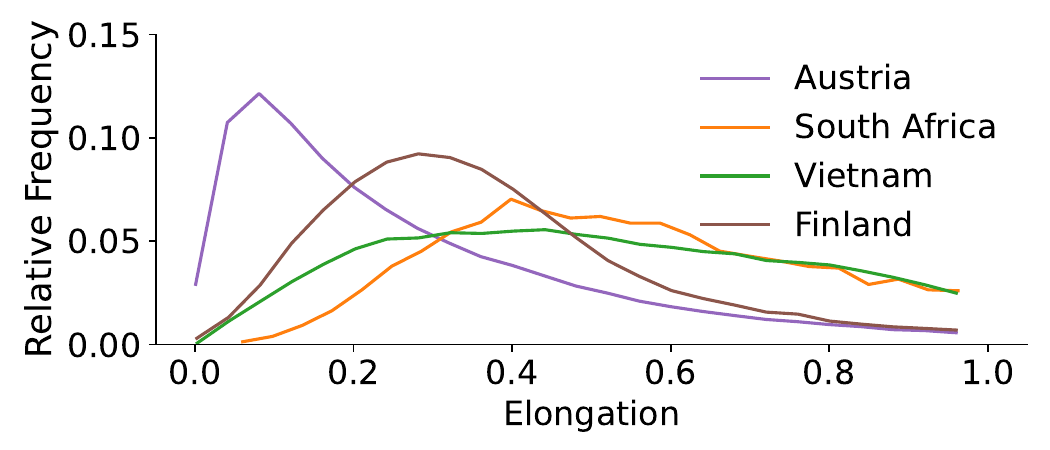}
    \caption{Elongation of field boundaries among different countries within the \ftw dataset.}
    \label{fig:elongation}
\end{figure}

Figure~\ref{fig:elongation} shows the distribution of field polygon elongation across four countries. To compute elongation, we first compute a minimum bounding rectangle for each field polygon. The elongation is then computed as the ratio of height (i.e., short-side length) to width (i.e., long-side length) of the minimum bounding rectangle, resulting in a value between 0 and 1. This shows, for example, that Austria has many long-narrow fields while the fields in Vietnam and South Africa are typically less elongated. 

\begin{figure}[ht]
    \centering
    \includegraphics[width=\columnwidth]{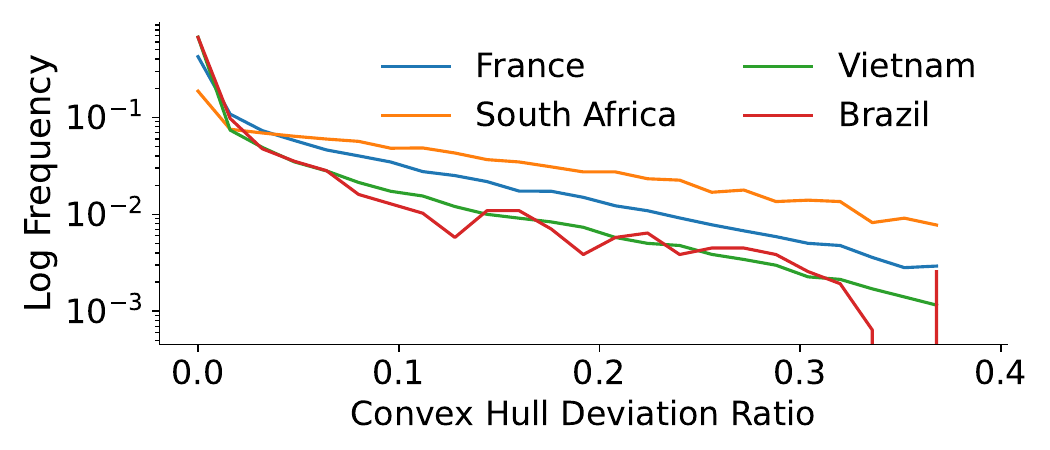}
    \caption{The Convex Hull Deviation Ratio among different countries within the \ftw dataset.}
    \label{fig:ch_dev_ratio}
\end{figure}

Figure~\ref{fig:ch_dev_ratio} shows distributions of the Convex Hull Deviation Ratio for different countries within the \ftw dataset. Let $f$ be the area of the field polygon and $c$ be the area of its convex hull. The Convex Hull Deviation Ratio is defined as $\frac{c - f}{c}$. This ratio is zero when the field is convex and increasingly close to one for highly non-convex field polygons. This shows that South Africa has heavy tails for the distribution, reflecting the relatively high number of highly non-convex field polygons, especially when compared to Brazil and Vietnam.

\begin{figure*}[ht]
    \centering
    \includegraphics[width=1\linewidth]{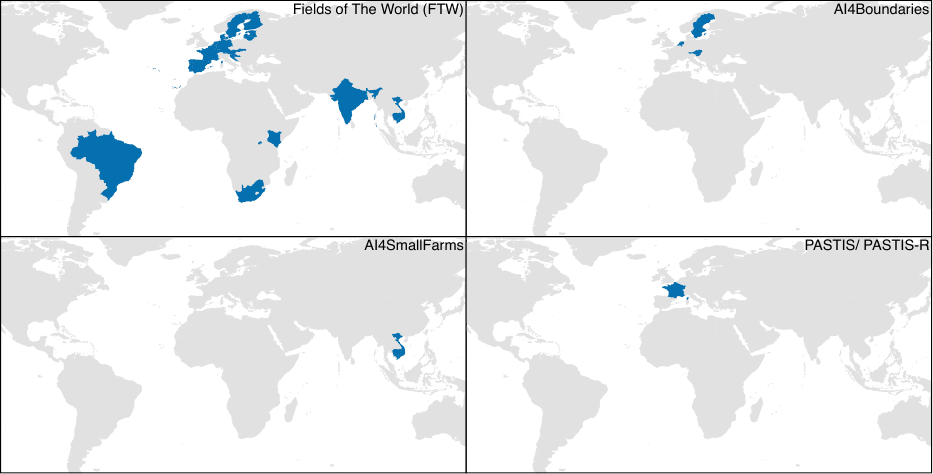}
    \caption{Geographical distribution and comparison of \ftw with other field boundary datasets. }
    \label{fig:dataset-dist}
\end{figure*}

Figure~\ref{fig:dataset-dist} presents a comparative analysis of the geographical coverage of various field boundary datasets.
AI4Boundaries and PASTIS/PASTIS-R datasets primarily cover European countries, while AI4SmallFarms focuses on two Asian countries. In contrast, the \ftw dataset spans multiple continents, including South America, Europe, Africa, and Asia.

Table~\ref{tab:climatological_diversity} compares the \ftw dataset with other field boundary datasets across current Köppen climate zones of the world \cite{beck2018present}. It shows that the AI4SmallFarms dataset exists only in a single climate zone, the equatorial savannah with dry winter, while the AI4Boundaries dataset spans 9 different climate zones, including two unique zones: Polar tundra and Warm temperate fully humid with a cool summer. The \ftw dataset is the most diverse among these, covering 17 different climate zones, including 7 unique zones where no other dataset is present.

\begin{figure*}[ht]
    \centering
    \includegraphics[width=1\linewidth]{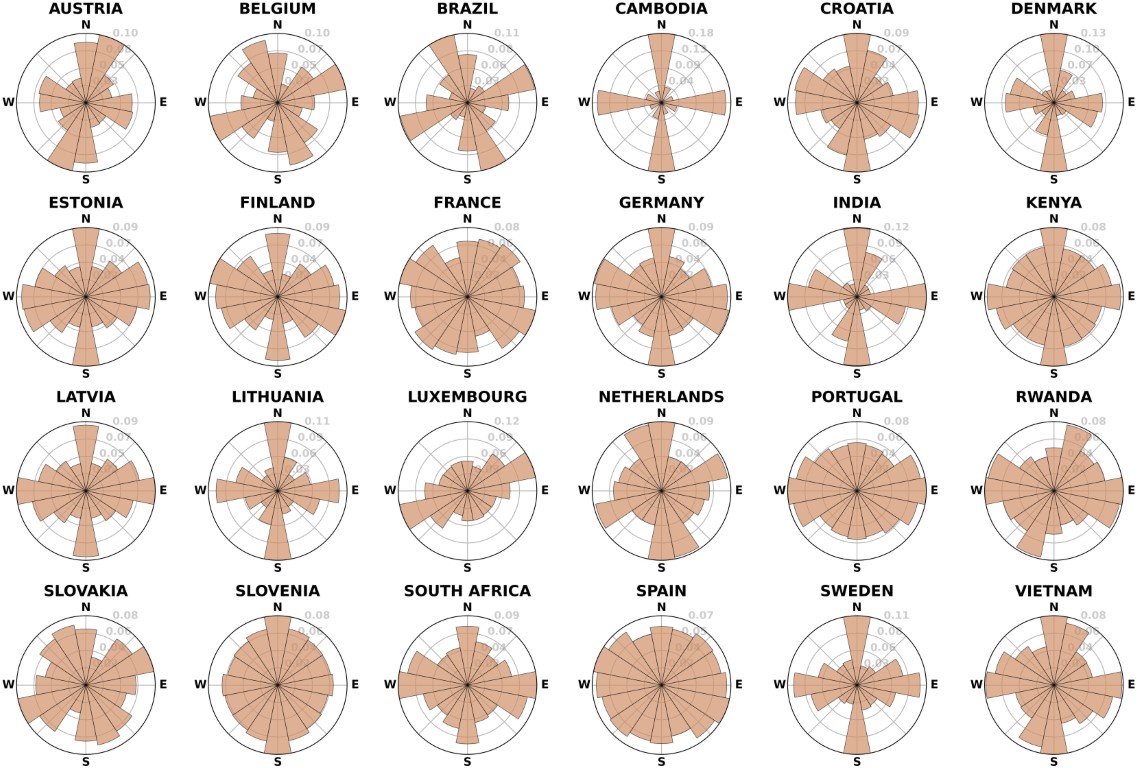}
    \caption{Field orientation histograms of all countries in the Fields of The World (\ftw) dataset.}
    \label{fig:oreint-all}
\end{figure*}

Figure~\ref{fig:oreint-all} shows the distribution of field orientations across all \ftw countries. Most countries exhibit diverse field orientations, while a few, such as Austria, Denmark, and Slovenia, have predominantly north-south orientations, and others, like Luxembourg, Portugal, and South Africa, have predominantly east-west orientations.

\begin{figure*}[ht]
    \centering
    \includegraphics[width=1\linewidth]{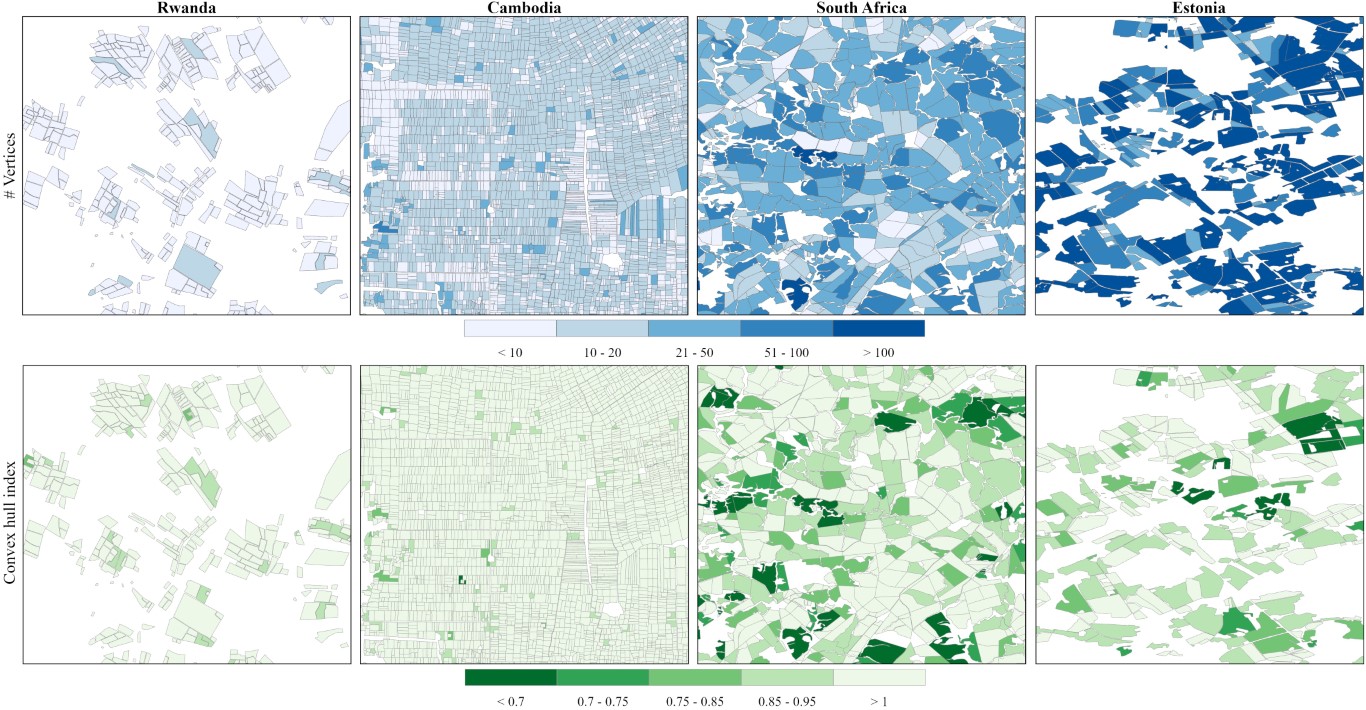}
    \caption{Visualization of the number of Field polygon vertices (above) and convex hull index for selected countries of the \ftw dataset.}
    \label{fig:morpho}
\end{figure*}

Figure~\ref{fig:morpho} shows the Convex Hull Index distributions for selected countries within the \ftw dataset. Let $p$ be the perimeter of the original field polygon, and $p_c$ be the perimeter of its convex hull; the Convex Hull Index is defined as $\frac{p_c}{p}$. This ratio provides insight into the complexity of a field's boundary, where values close to 1 indicate that field polygons are nearly convex, and values significantly less than 1 suggest that the polygons are non-convex with more complex boundaries. The figure shows that the field boundaries in South Africa and Estonia are more non-convex/ complex than those in Rwanda and Cambodia.

\begin{table*}[ht]
\caption{Climatological diversity and comparison of Fields of The World (\ftw) dataset with previous field boundary datasets, showing the total number of field polygons in each climatic zone.}
\resizebox{\textwidth}{!}{%
\begin{tabular}{@{}lllllllllll@{}}
\toprule
\textbf{Köppen Climate zone} & Fields of The World (\ftw) & AI4Boundaries & AI4SmallFarms \\ \midrule
Polar tundra            & 0 & 8,273 & 0 \\
Warm temperate fully humid with cool summer            & 0 & 2,293 & 0 \\
Warm temperate with dry winter and warm summer           & 62 & 0 & 0 \\
Arid Steppe cold     & 91 & 11,709 & 0 \\
Equatorial rainforest, fully humid      & 100 & 0 & 0 \\
Arid desert hot       & 596 & 0 & 0 \\
Equatorial savannah with dry summer       & 725 & 0 & 0 \\
Warm temperate with dry, warm summer     & 2,686 & 3,980 & 0 \\
Arid Steppe hot & 3,344 & 0 & 0 \\
Warm temperate with dry, hot summer & 8,681 & 129,618 & 0 \\
Equatorial monsoon & 25,587 & 0 & 0 \\
Snow fully humid cool summer & 38,870 & 35,109 & 0 \\
Warm temperate with dry winter and hot summer & 51,140 & 0 & 0 \\
Warm temperate fully humid with hot summer & 114,745 & 12,508 & 0 \\
Snow fully humid warm summer & 168,334 & 65,721 & 0 \\
Equatorial savannah with dry winter & 370,259 & 0 & 126,672 \\ 
Warm temperate fully humid with warm summer & 842,158 & 800,012 & 0 \\ \midrule
Total & 1,627,378 & 1,069,223 & 126,672 \\ \bottomrule
\end{tabular}%
}
\label{tab:climatological_diversity}
\end{table*}

\subsection{Prediction heatmaps}

\begin{figure*}[ht]
    \centering
    \includegraphics[width=0.95\linewidth]{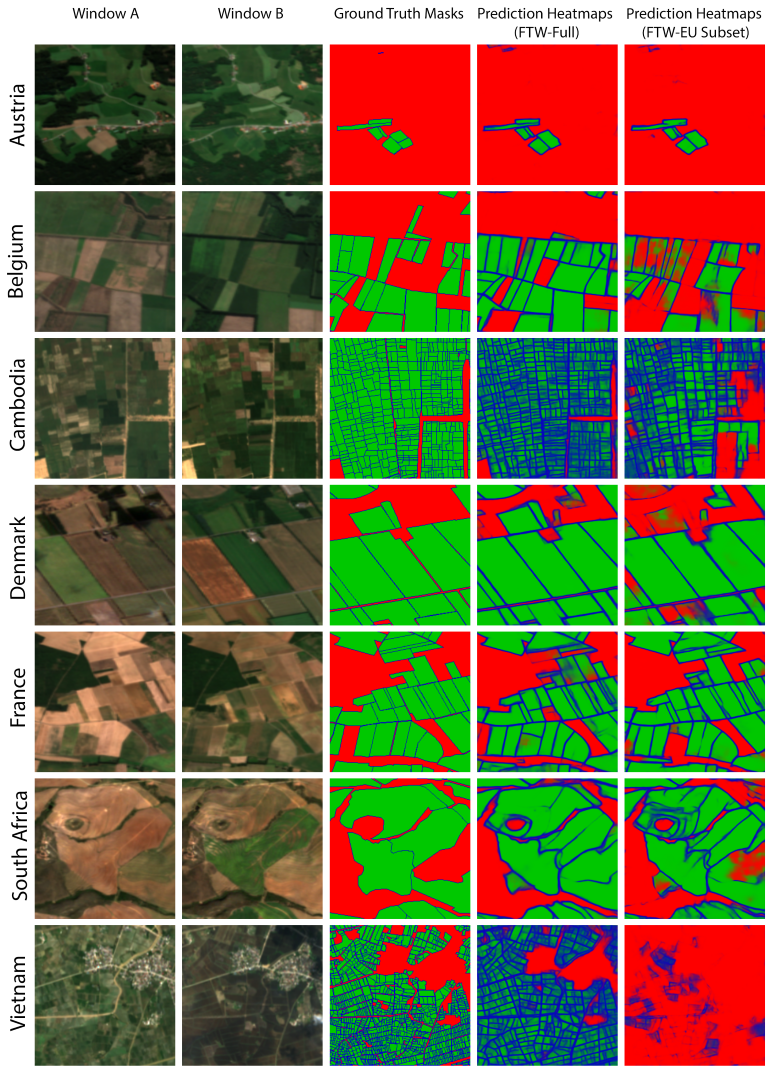}
    \caption{Prediction Heatmaps from Presence/Absence countries (R: Background, G: Fields, B: Boundaries)}
    \label{fig:heatmap_visualization_predictions_pa}
\end{figure*}

\begin{figure*}[ht]
    \centering
    \includegraphics[width=0.95\linewidth]{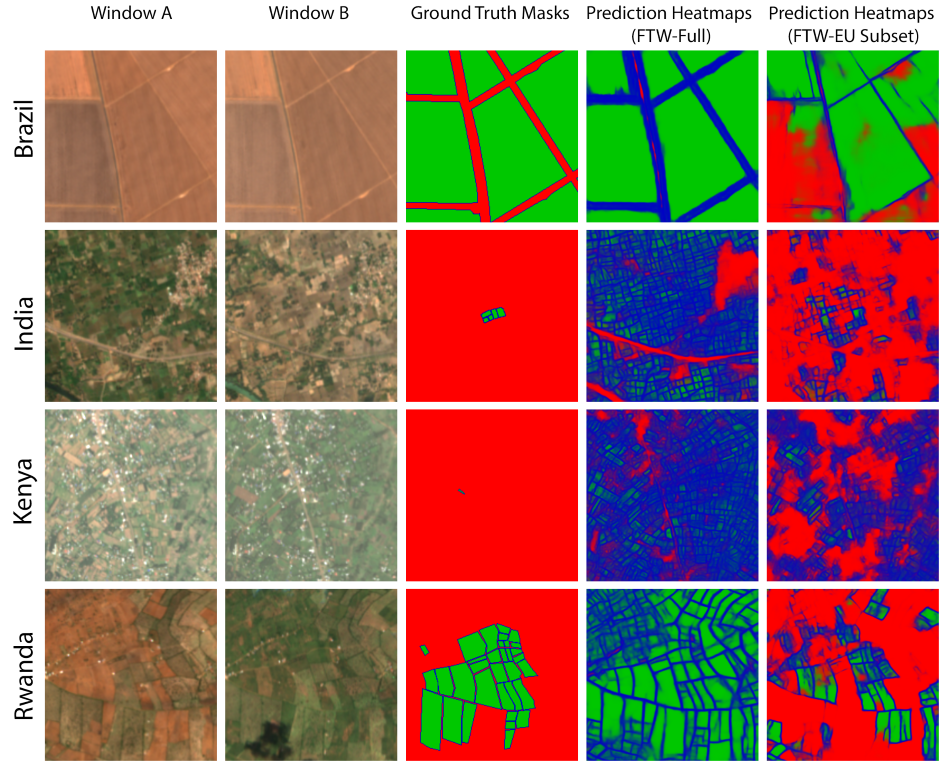}
    \caption{Prediction samples from Presence-Only countries (R: Background, G: Fields, B: Boundaries)}
    \label{fig:heatmap_visualization_predictions_po}
\end{figure*}

In Figure~\ref{fig:heatmap_visualization_predictions_pa} and Figure~\ref{fig:heatmap_visualization_predictions_po}, we visualize the class prediction heatmaps for the U-Net with EfficientNet-b3 backbone trained on the full \ftw dataset (\ftw-Full) with 3-class masks (ignore presence-only background). We also visualize the heatmaps for the same model trained on the subset of European countries (\ftw-EU) that are in AI4Boundaries (Austria, Spain, France, Luxembourg, Netherlands, Slovenia, and Sweden). 

Figures \ref{fig:heatmap_visualization_predictions_pa} and \ref{fig:heatmap_visualization_predictions_po} illustrate these predictions for a sample of 8 countries, representing both Presence/Absence and Presence-only regions (respectively).
The heatmaps use three classes for prediction: Red for the Background class, Green for the Field Extent (Interior) class, and Blue for the Boundary class.

Our results show that the \ftw-Full predictions are more aligned with the ground truth for both European and non-European countries. Notably, the \ftw-Full model provides more accurate boundary predictions (blue channel) in countries with smaller fields, such as Cambodia, Vietnam, India, and Kenya.

In contrast, the \ftw-EU model struggles with accurate predictions in Presence-only regions, particularly for the Field Extent and Boundary classes. However, in some cases, such as France, the \ftw-EU confidently predicts the Field Extent class, sometimes more accurately aligning with the ground truth than \ftw-Full.

These visualizations help illustrate how using different datasets for training affects the model's predictions in different regions. By comparing the \ftw-Full with the \ftw-EU model heatmaps, we can see that the \ftw-Full heatmaps align better with the ground truth masks across diverse field patterns globally.  

\subsection{Experiments}

\subsubsection{Model architecture experiment results}

We evaluated two semantic segmentation model architectures, U-Net \cite{ronneberger2015u} and DeepLabv3+ \cite{chen2018encoder}, with five different backbones: ResNet-18, ResNet-50, ResNeXt-50, EfficientNet-b3, and EfficientNet-b4. We report performance in Table~\ref{tab:exp1-architecture} for Slovenia, France, and South Africa. U-Nets performed slightly better than DeepLabv3+ models, and U-Nets with EfficientNet backbones performed best. 

\begin{table*}[ht]
\caption{Performance metrics for various model architectures in Slovenia (SVN), France (FRA), and South Africa (ZAF).}
\resizebox{\textwidth}{!}{%
\begin{tabular}{@{}llllllllllllllll@{}}
\toprule
\multicolumn{1}{c}{\multirow{2}{*}{\textbf{Architecture + backbone}}} & \multicolumn{3}{c}{Pixel IoU} & \multicolumn{3}{c}{Pixel precision} & \multicolumn{3}{c}{Pixel recall} & \multicolumn{3}{c}{Object precision} & \multicolumn{3}{c}{Object recall} \\ \cmidrule(l){2-16} 
\multicolumn{1}{c}{} & SVN & FRA & ZAF & SVN & FRA & ZAF & SVN & FRA & ZAF & SVN & FRA & ZAF & SVN & FRA & ZAF \\ \midrule
U-net + ResNet-18 & 0.53 & 0.77 & 0.79 & 0.89 & 0.88 & \textbf{0.90} & 0.57 & 0.86 & 0.87 & 0.24 & 0.49 & 0.48 & 0.15 & 0.54 & 0.47 \\
U-net + ResNet-50 & 0.55 & 0.78 & 0.79 & 0.89 & 0.89 & \textbf{0.90} & 0.59 & 0.87 & 0.87 & 0.26 & 0.52 & 0.47 & 0.16 & 0.54 & 0.50 \\
U-net + ResNeXt-50 & 0.56 & \textbf{0.79} & \textbf{0.80} & 0.90 & 0.89 & 0.89 & 0.60 & \textbf{0.88} & 0.88 & 0.29 & 0.54 & 0.49 & 0.17 & 0.56 & 0.49 \\
U-net + EfficientNet-b3 & \textbf{0.59} & \textbf{0.79} & \textbf{0.80} & \textbf{0.91} & 0.89 & 0.89 & \textbf{0.63} & 0.87 & 0.88 & \textbf{0.31} & 0.54 & \textbf{0.53} & \textbf{0.19} & 0.58 & \textbf{0.56} \\
U-net + EfficientNet-b4 & \textbf{0.59} & \textbf{0.79} & 0.79 & 0.90 & 0.89 & 0.88 & \textbf{0.63} & \textbf{0.88} & \textbf{0.89} & \textbf{0.31} & \textbf{0.55} & 0.52 & \textbf{0.19} & \textbf{0.59} & 0.55 \\
DeepLabv3+ + ResNet-18 & 0.43 & 0.75 & 0.77 & 0.87 & 0.88 & 0.89 & 0.46 & 0.84 & 0.85 & 0.20 & 0.47 & 0.45 & 0.08 & 0.47 & 0.43 \\
DeepLabv3+ + ResNet-50 & 0.47 & 0.76 & 0.79 & 0.89 & 0.89 & \textbf{0.90} & 0.50 & 0.84 & 0.87 & 0.22 & 0.49 & 0.48 & 0.10 & 0.49 & 0.44 \\
DeepLabv3+ + ResNeXt-50 & 0.48 & 0.77 & 0.79 & 0.88 & 0.89 & 0.89 & 0.51 & 0.85 & 0.87 & 0.23 & 0.50 & 0.48 & 0.10 & 0.49 & 0.44 \\
DeepLabv3+ + EfficientNet-b3 & 0.50 & 0.77 & 0.78 & 0.90 & 0.89 & 0.89 & 0.53 & 0.84 & 0.87 & 0.21 & 0.49 & 0.47 & 0.11 & 0.50 & 0.47 \\
DeepLabv3+ + EfficientNet-b4 & 0.50 & 0.77 & 0.79 & \textbf{0.91} & \textbf{0.90} & 0.89 & 0.53 & 0.85 & 0.88 & 0.24 & 0.51 & 0.47 & 0.12 & 0.50 & 0.47 \\ 
\bottomrule
\end{tabular}%
}
\label{tab:exp1-architecture}
\end{table*}

\subsubsection{Per-country experiment results}

In the experiment results in Tables 3 and 4 of the main paper, we reported results for a subset of test countries in \ftw (Slovenia, France, and South Africa). We provide the full results of those experiments for all test countries in Tables~\ref{tab:exp1-masktype-full}-\ref{tab:exp1-masktype-full-lz} and Tables~\ref{tab:exp1-channels-fulla}-\ref{tab:exp1-channels-fullc} (respectively) of the supplement.

\begin{table*}[ht]
\caption{Performance metrics for different target mask formats in all test countries (names starting with A-K). 
We compared 2-class field extent and 3-class masks with or without ignoring background (bg) pixels for presence-only samples. We only report recall metrics for presence-only countries.}
\label{tab:exp1-masktype-full}
\resizebox{\textwidth}{!}{%
\begin{tabular}{@{}ccccccc@{}}
\toprule
\textbf{Test country} & \textbf{Mask type} & \textbf{Pixel IoU} & \textbf{Pixel precision} & \textbf{Pixel recall} & \textbf{Object precision} & \textbf{Object recall} \\ \midrule
\multirow{4}{*}{Austria} & 2-class & 0.77 & 0.84 & 0.91 & 0.37 & 0.11 \\
 & 2-class (ignore presence-only bg) & 0.77 & 0.83 & 0.92 & 0.36 & 0.11 \\
 & 3-class & 0.71 & 0.91 & 0.76 & 0.44 & 0.38 \\
 & 3-class (ignore presence-only bg) & 0.70 & 0.90 & 0.76 & 0.44 & 0.39 \\ \midrule
\multirow{4}{*}{Belgium} & 2-class & 0.80 & 0.86 & 0.92 & 0.48 & 0.24 \\
 & 2-class (ignore presence-only bg) & 0.80 & 0.86 & 0.93 & 0.46 & 0.23 \\
 & 3-class & 0.75 & 0.93 & 0.79 & 0.56 & 0.58 \\
 & 3-class (ignore presence-only bg) & 0.75 & 0.92 & 0.80 & 0.57 & 0.58 \\ \midrule
\multirow{4}{*}{Brazil} & 2-class & - & - & 1.00 & 0.16 & 0.11 \\
 & 2-class (ignore presence-only bg) & - & - & 1.00 & - & 0.12 \\
 & 3-class & - & - & 0.96 & - & 0.60 \\
 & 3-class (ignore presence-only bg) & - & - & 0.96 & - & 0.58 \\ \midrule
\multirow{4}{*}{Cambodia} & 2-class & 0.76 & 0.78 & 0.97 & 0.06 & 0.00 \\
 & 2-class (ignore presence-only bg) & 0.76 & 0.78 & 0.97 & 0.06 & 0.00 \\
 & 3-class & 0.40 & 0.95 & 0.40 & 0.22 & 0.17 \\
 & 3-class (ignore presence-only bg) & 0.43 & 0.95 & 0.44 & 0.26 & 0.20 \\ \midrule
\multirow{4}{*}{Corsica} & 2-class & 0.49 & 0.69 & 0.63 & 0.16 & 0.07 \\
 & 2-class (ignore presence-only bg) & 0.49 & 0.68 & 0.64 & 0.17 & 0.09 \\
 & 3-class & 0.45 & 0.76 & 0.53 & 0.21 & 0.16 \\
 & 3-class (ignore presence-only bg) & 0.48 & 0.79 & 0.55 & 0.21 & 0.17 \\ \midrule
\multirow{4}{*}{Croatia} & 2-class & 0.76 & 0.80 & 0.94 & 0.26 & 0.08 \\
 & 2-class (ignore presence-only bg) & 0.75 & 0.79 & 0.93 & 0.24 & 0.08 \\
 & 3-class & 0.67 & 0.89 & 0.73 & 0.25 & 0.33 \\
 & 3-class (ignore presence-only bg) & 0.68 & 0.89 & 0.74 & 0.25 & 0.34 \\ \midrule
\multirow{4}{*}{Denmark} & 2-class & 0.84 & 0.89 & 0.94 & 0.41 & 0.28 \\
 & 2-class (ignore presence-only bg) & 0.84 & 0.89 & 0.94 & 0.38 & 0.25 \\
 & 3-class & 0.83 & 0.93 & 0.88 & 0.45 & 0.61 \\
 & 3-class (ignore presence-only bg) & 0.83 & 0.93 & 0.88 & 0.45 & 0.60 \\ \midrule
\multirow{4}{*}{Estonia} & 2-class & 0.81 & 0.88 & 0.91 & 0.47 & 0.29 \\
 & 2-class (ignore presence-only bg) & 0.81 & 0.88 & 0.91 & 0.48 & 0.29 \\
 & 3-class & 0.80 & 0.92 & 0.86 & 0.46 & 0.42 \\
 & 3-class (ignore presence-only bg) & 0.79 & 0.91 & 0.86 & 0.47 & 0.43 \\ \midrule
\multirow{4}{*}{Finland} & 2-class & 0.87 & 0.90 & 0.96 & 0.42 & 0.18 \\
 & 2-class (ignore presence-only bg) & 0.87 & 0.90 & 0.96 & 0.42 & 0.18 \\
 & 3-class & 0.83 & 0.96 & 0.86 & 0.54 & 0.56 \\
 & 3-class (ignore presence-only bg) & 0.83 & 0.96 & 0.87 & 0.55 & 0.57 \\ \midrule
\multirow{4}{*}{France} & 2-class & 0.82 & 0.85 & 0.95 & 0.39 & 0.16 \\
 & 2-class (ignore presence-only bg) & 0.82 & 0.86 & 0.95 & 0.38 & 0.15 \\
 & 3-class & 0.79 & 0.89 & 0.87 & 0.53 & 0.57 \\
 & 3-class (ignore presence-only bg) & 0.79 & 0.89 & 0.88 & 0.55 & 0.58 \\ \midrule
\multirow{4}{*}{Germany} & 2-class & 0.80 & 0.84 & 0.94 & 0.40 & 0.19 \\
 & 2-class (ignore presence-only bg) & 0.79 & 0.84 & 0.93 & 0.37 & 0.18 \\
 & 3-class & 0.79 & 0.87 & 0.89 & 0.42 & 0.40 \\
 & 3-class (ignore presence-only bg) & 0.79 & 0.87 & 0.90 & 0.43 & 0.42 \\ \midrule
\multirow{4}{*}{India} & 2-class & - & - & 0.99 & - & 0.00 \\
 & 2-class (ignore presence-only bg) & - & - & 0.99 & - & 0.00 \\
 & 3-class & - & - & 0.23 & - & 0.05 \\
 & 3-class (ignore presence-only bg) & - & - & 0.22 & - & 0.06 \\ \midrule
\multirow{4}{*}{Kenya} & 2-class & - & - & 0.95 & - & 0.00 \\
 & 2-class (ignore presence-only bg) & - & - & 0.97 & - & 0.00 \\
 & 3-class & - & - & 0.47 & - & 0.08 \\
 & 3-class (ignore presence-only bg) & - & - & 0.49 & - & 0.10 \\ \bottomrule
\end{tabular}%
}
\end{table*}

\begin{table*}[ht]
\caption{Performance metrics for different target mask formats in test countries (names starting with L-Z). 
We compared 2-class field extent and 3-class masks with or without ignoring background (bg) pixels for presence-only samples. We only report recall metrics for presence-only countries.}
\label{tab:exp1-masktype-full-lz}
\resizebox{\textwidth}{!}{%
\begin{tabular}{@{}ccccccc@{}}
\toprule
\textbf{Test country} & \textbf{Mask type} & \textbf{Pixel IoU} & \textbf{Pixel precision} & \textbf{Pixel recall} & \textbf{Object precision} & \textbf{Object recall} \\ \midrule
\multirow{4}{*}{Latvia} & 2-class & 0.84 & 0.90 & 0.92 & 0.43 & 0.26 \\
 & 2-class (ignore presence-only bg) & 0.84 & 0.90 & 0.92 & 0.44 & 0.27 \\
 & 3-class & 0.81 & 0.94 & 0.85 & 0.43 & 0.45 \\
 & 3-class (ignore presence-only bg) & 0.81 & 0.94 & 0.86 & 0.44 & 0.45 \\ \midrule
\multirow{4}{*}{Lithuania} & 2-class & 0.78 & 0.83 & 0.93 & 0.38 & 0.19 \\
 & 2-class (ignore presence-only bg) & 0.77 & 0.82 & 0.93 & 0.39 & 0.18 \\
 & 3-class & 0.74 & 0.88 & 0.82 & 0.37 & 0.41 \\
 & 3-class (ignore presence-only bg) & 0.74 & 0.88 & 0.83 & 0.37 & 0.41 \\ \midrule
\multirow{4}{*}{Luxembourg} & 2-class & 0.85 & 0.88 & 0.97 & 0.25 & 0.05 \\
 & 2-class (ignore presence-only bg) & 0.86 & 0.88 & 0.97 & 0.24 & 0.05 \\
 & 3-class & 0.79 & 0.97 & 0.81 & 0.47 & 0.52 \\
 & 3-class (ignore presence-only bg) & 0.79 & 0.96 & 0.82 & 0.47 & 0.51 \\ \midrule
\multirow{4}{*}{Netherlands} & 2-class & 0.79 & 0.86 & 0.91 & 0.49 & 0.25 \\
 & 2-class (ignore presence-only bg) & 0.79 & 0.86 & 0.91 & 0.48 & 0.24 \\
 & 3-class & 0.75 & 0.92 & 0.80 & 0.51 & 0.45 \\
 & 3-class (ignore presence-only bg) & 0.75 & 0.92 & 0.80 & 0.53 & 0.45 \\ \midrule
\multirow{4}{*}{Portugal} & 2-class & 0.29 & 0.41 & 0.51 & 0.03 & 0.01 \\
 & 2-class (ignore presence-only bg) & 0.31 & 0.65 & 0.37 & 0.04 & 0.01 \\
 & 3-class & 0.12 & 0.79 & 0.12 & 0.05 & 0.02 \\
 & 3-class (ignore presence-only bg) & 0.12 & 0.67 & 0.12 & 0.07 & 0.03 \\ \midrule
\multirow{4}{*}{Rwanda} & 2-class & - & - & 0.99 & - & 0.00 \\
 & 2-class (ignore presence-only bg) & - & - & 0.98 & - & 0.00 \\
 & 3-class & - & - & 0.55 & - & 0.26 \\
  & 3-class (ignore presence-only bg) & - & - & 0.57 & - & 0.30 \\ \midrule
\multirow{4}{*}{Slovakia} & 2-class & 0.93 & 0.95 & 0.98 & 0.59 & 0.40 \\
 & 2-class (ignore presence-only bg) & 0.93 & 0.95 & 0.98 & 0.59 & 0.40 \\
 & 3-class & 0.92 & 0.98 & 0.94 & 0.51 & 0.55 \\
 & 3-class (ignore presence-only bg) & 0.92 & 0.98 & 0.95 & 0.50 & 0.55 \\ \midrule
\multirow{4}{*}{Slovenia} & 2-class & 0.69 & 0.79 & 0.85 & 0.30 & 0.08 \\
 & 2-class (ignore presence-only bg) & 0.69 & 0.78 & 0.86 & 0.30 & 0.08 \\
 & 3-class & 0.58 & 0.90 & 0.62 & 0.31 & 0.19 \\
 & 3-class (ignore presence-only bg) & 0.59 & 0.90 & 0.63 & 0.33 & 0.20 \\ \midrule
\multirow{4}{*}{South Africa} & 2-class & 0.82 & 0.85 & 0.96 & 0.44 & 0.20 \\
 & 2-class (ignore presence-only bg) & 0.81 & 0.84 & 0.96 & 0.44 & 0.19 \\
 & 3-class & 0.80 & 0.90 & 0.88 & 0.51 & 0.55 \\
 & 3-class (ignore presence-only bg) & 0.79 & 0.89 & 0.88 & 0.51 & 0.55 \\ \midrule
\multirow{4}{*}{Spain} & 2-class & 0.84 & 0.87 & 0.95 & 0.33 & 0.06 \\
 & 2-class (ignore presence-only bg) & 0.84 & 0.87 & 0.95 & 0.32 & 0.05 \\
 & 3-class & 0.73 & 0.96 & 0.75 & 0.33 & 0.19 \\
 & 3-class (ignore presence-only bg) & 0.74 & 0.96 & 0.76 & 0.36 & 0.20 \\ \midrule
\multirow{4}{*}{Sweden} & 2-class & 0.83 & 0.88 & 0.93 & 0.37 & 0.20 \\
 & 2-class (ignore presence-only bg) & 0.83 & 0.88 & 0.93 & 0.34 & 0.19 \\
 & 3-class & 0.81 & 0.94 & 0.85 & 0.41 & 0.51 \\
 & 3-class (ignore presence-only bg) & 0.81 & 0.94 & 0.85 & 0.40 & 0.51 \\ \midrule
\multirow{4}{*}{Vietnam} & 2-class & 0.67 & 0.70 & 0.94 & 0.09 & 0.01 \\
 & 2-class (ignore presence-only bg) & 0.67 & 0.70 & 0.95 & 0.10 & 0.01 \\
 & 3-class & 0.46 & 0.89 & 0.49 & 0.18 & 0.13 \\
 & 3-class (ignore presence-only bg) & 0.47 & 0.89 & 0.49 & 0.18 & 0.13 \\ \bottomrule
\end{tabular}%
}
\end{table*}

\begin{table*}[ht]
\caption{Performance for input channel ablations in all test countries (names starting with A-G): multispectral (RGB-NIR vs. RGB only) and multi-temporal (Window A and Window B) input channels.}
\label{tab:exp1-channels-fulla}
\resizebox{\textwidth}{!}{%
\begin{tabular}{@{}ccccccc@{}}
\toprule
\textbf{Test country} & \textbf{Channels} & \textbf{Pixel IoU} & \textbf{Pixel precision} & \textbf{Pixel recall} & \textbf{Object precision} & \textbf{Object recall} \\ \midrule
\multirow{5}{*}{Austria} & Stacked Windows A and B (RGB only) & 0.70 & 0.90 & 0.76 & 0.42 & 0.37 \\
 & Stacked Windows A and B & 0.71 & 0.90 & 0.77 & 0.44 & 0.39 \\
 & Mean of Windows A and B & 0.66 & 0.87 & 0.73 & 0.39 & 0.34 \\
 & Window A only & 0.67 & 0.90 & 0.73 & 0.37 & 0.34 \\
 & Window B only & 0.68 & 0.88 & 0.74 & 0.40 & 0.35 \\ \midrule
\multirow{5}{*}{Belgium} & Stacked Windows A and B (RGB only) & 0.72 & 0.91 & 0.77 & 0.51 & 0.55 \\
 & Stacked Windows A and B & 0.75 & 0.93 & 0.80 & 0.58 & 0.58 \\
 & Mean of Windows A and B & 0.69 & 0.89 & 0.75 & 0.48 & 0.53 \\
 & Window A only & 0.73 & 0.93 & 0.77 & 0.51 & 0.53 \\
 & Window B only & 0.65 & 0.84 & 0.74 & 0.42 & 0.51 \\ \midrule
\multirow{5}{*}{Brazil} & Stacked Windows A and B (RGB only) & - & - & 0.96 & - & 0.59 \\
 & Stacked Windows A and B & - & - & 0.96 & - & 0.58 \\
 & Mean of Windows A and B & - & - & 0.96 & - & 0.60 \\
 & Window A only & - & - & 0.96 & - & 0.56 \\
 & Window B only & - & - & 0.96 & - & 0.60 \\ \midrule
\multirow{5}{*}{Cambodia} & Stacked Windows A and B (RGB only) & 0.35 & 0.94 & 0.36 & 0.20 & 0.15 \\
 & Stacked Windows A and B & 0.39 & 0.95 & 0.39 & 0.22 & 0.17 \\
 & Mean of Windows A and B & 0.33 & 0.95 & 0.33 & 0.17 & 0.13 \\
 & Window A only & 0.35 & 0.94 & 0.36 & 0.15 & 0.13 \\
 & Window B only & 0.38 & 0.94 & 0.39 & 0.20 & 0.15 \\ \midrule
\multirow{5}{*}{Corsica} & Stacked Windows A and B (RGB only) & 0.45 & 0.79 & 0.51 & 0.24 & 0.16 \\
 & Stacked Windows A and B & 0.47 & 0.80 & 0.54 & 0.26 & 0.17 \\
 & Mean of Windows A and B & 0.42 & 0.73 & 0.50 & 0.18 & 0.14 \\
 & Window A only & 0.45 & 0.81 & 0.50 & 0.23 & 0.17 \\
 & Window B only & 0.42 & 0.75 & 0.49 & 0.19 & 0.13 \\ \midrule
\multirow{5}{*}{Croatia} & Stacked Windows A and B (RGB only) & 0.66 & 0.88 & 0.72 & 0.23 & 0.32 \\
 & Stacked Windows A and B & 0.67 & 0.89 & 0.73 & 0.25 & 0.33 \\
 & Mean of Windows A and B & 0.64 & 0.88 & 0.71 & 0.24 & 0.29 \\
 & Window A only & 0.65 & 0.87 & 0.72 & 0.22 & 0.31 \\
 & Window B only & 0.66 & 0.89 & 0.72 & 0.22 & 0.29 \\ \midrule
\multirow{5}{*}{Denmark} & Stacked Windows A and B (RGB only) & 0.82 & 0.93 & 0.87 & 0.43 & 0.59 \\
 & Stacked Windows A and B & 0.83 & 0.93 & 0.88 & 0.46 & 0.60 \\
 & Mean of Windows A and B & 0.82 & 0.93 & 0.88 & 0.43 & 0.58 \\
 & Window A only & 0.82 & 0.93 & 0.88 & 0.44 & 0.60 \\
 & Window B only & 0.81 & 0.93 & 0.87 & 0.43 & 0.58 \\ \midrule
\multirow{5}{*}{Estonia} & Stacked Windows A and B (RGB only) & 0.78 & 0.91 & 0.85 & 0.45 & 0.41 \\
 & Stacked Windows A and B & 0.80 & 0.92 & 0.86 & 0.49 & 0.42 \\
 & Mean of Windows A and B & 0.78 & 0.90 & 0.85 & 0.45 & 0.41 \\
 & Window A only & 0.79 & 0.93 & 0.85 & 0.48 & 0.41 \\
 & Window B only & 0.77 & 0.90 & 0.84 & 0.42 & 0.40 \\ \midrule
\multirow{5}{*}{Finland} & Stacked Windows A and B (RGB only) & 0.82 & 0.95 & 0.85 & 0.52 & 0.54 \\
 & Stacked Windows A and B & 0.83 & 0.96 & 0.86 & 0.56 & 0.56 \\
 & Mean of Windows A and B & 0.81 & 0.95 & 0.84 & 0.52 & 0.52 \\
 & Window A only & 0.82 & 0.95 & 0.85 & 0.51 & 0.54 \\
 & Window B only & 0.80 & 0.95 & 0.84 & 0.51 & 0.52 \\ \midrule
\multirow{5}{*}{France} & Stacked Windows A and B (RGB only) & 0.78 & 0.89 & 0.86 & 0.51 & 0.56 \\
 & Stacked Windows A and B & 0.79 & 0.89 & 0.87 & 0.54 & 0.58 \\
 & Mean of Windows A and B & 0.77 & 0.89 & 0.86 & 0.50 & 0.55 \\
 & Window A only & 0.77 & 0.88 & 0.86 & 0.47 & 0.54 \\
 & Window B only & 0.78 & 0.89 & 0.86 & 0.49 & 0.54 \\ \midrule
\multirow{5}{*}{Germany} & Stacked Windows A and B (RGB only) & 0.78 & 0.87 & 0.88 & 0.41 & 0.41 \\
 & Stacked Windows A and B & 0.79 & 0.87 & 0.89 & 0.43 & 0.41 \\
 & Mean of Windows A and B & 0.77 & 0.87 & 0.87 & 0.41 & 0.38 \\
 & Window A only & 0.77 & 0.87 & 0.88 & 0.37 & 0.37 \\
 & Window B only & 0.78 & 0.87 & 0.89 & 0.40 & 0.41 \\
 \bottomrule
\end{tabular}%
}
\end{table*}

\begin{table*}[ht]
\caption{Performance for input channel ablations in all test countries (names starting with H-R): multispectral (RGB-NIR vs. RGB only) and multi-temporal (Window A and Window B) input channels.}
\label{tab:exp1-channels-fullb}
\resizebox{\textwidth}{!}{%
\begin{tabular}{@{}ccccccc@{}}
\toprule
\textbf{Test country} & \textbf{Channels} & \textbf{Pixel IoU} & \textbf{Pixel precision} & \textbf{Pixel recall} & \textbf{Object precision} & \textbf{Object recall} \\ \midrule
\multirow{5}{*}{India} & Stacked Windows A and B (RGB only) & - & - & 0.22 & - & 0.04 \\
 & Stacked Windows A and B & - & - & 0.20 & - & 0.05 \\
 & Mean of Windows A and B & - & - & 0.20 & - & 0.04 \\
 & Window A only & - & - & 0.29 & - & 0.06 \\
 & Window B only & - & - & 0.17 & - & 0.02 \\ \midrule
\multirow{5}{*}{Kenya} & Stacked Windows A and B (RGB only) & - & - & 0.48 & - & 0.10 \\
 & Stacked Windows A and B & - & - & 0.47 & - & 0.10 \\
 & Mean of Windows A and B & - & - & 0.46 & - & 0.09 \\
 & Window A only & - & - & 0.40 & - & 0.06 \\
 & Window B only & - & - & 0.47 & - & 0.08 \\ \midrule
\multirow{5}{*}{Latvia} & Stacked Windows A and B (RGB only) & 0.80 & 0.94 & 0.85 & 0.43 & 0.44 \\
 & Stacked Windows A and B & 0.81 & 0.94 & 0.86 & 0.45 & 0.45 \\
 & Mean of Windows A and B & 0.79 & 0.92 & 0.84 & 0.41 & 0.43 \\
 & Window A only & 0.80 & 0.93 & 0.85 & 0.40 & 0.43 \\
 & Window B only & 0.78 & 0.92 & 0.83 & 0.39 & 0.41 \\ \midrule
\multirow{5}{*}{Lithuania} & Stacked Windows A and B (RGB only) & 0.71 & 0.86 & 0.80 & 0.35 & 0.39 \\
 & Stacked Windows A and B & 0.73 & 0.88 & 0.82 & 0.37 & 0.41 \\
 & Mean of Windows A and B & 0.69 & 0.84 & 0.79 & 0.33 & 0.37 \\
 & Window A only & 0.72 & 0.87 & 0.81 & 0.34 & 0.37 \\
 & Window B only & 0.59 & 0.74 & 0.74 & 0.26 & 0.34 \\ \midrule
\multirow{5}{*}{Luxembourg} & Stacked Windows A and B (RGB only) & 0.78 & 0.96 & 0.81 & 0.46 & 0.51 \\
 & Stacked Windows A and B & 0.79 & 0.97 & 0.81 & 0.47 & 0.51 \\
 & Mean of Windows A and B & 0.77 & 0.97 & 0.80 & 0.44 & 0.50 \\
 & Window A only & 0.78 & 0.96 & 0.80 & 0.42 & 0.46 \\
 & Window B only & 0.77 & 0.96 & 0.80 & 0.43 & 0.49 \\ \midrule
\multirow{5}{*}{Netherlands} & Stacked Windows A and B (RGB only) & 0.71 & 0.91 & 0.77 & 0.47 & 0.43 \\
 & Stacked Windows A and B & 0.76 & 0.92 & 0.81 & 0.52 & 0.45 \\
 & Mean of Windows A and B & 0.70 & 0.90 & 0.76 & 0.44 & 0.42 \\
 & Window A only & 0.65 & 0.86 & 0.72 & 0.38 & 0.38 \\
 & Window B only & 0.72 & 0.91 & 0.78 & 0.46 & 0.42 \\ \midrule
\multirow{5}{*}{Portugal} & Stacked Windows A and B (RGB only) & 0.14 & 0.82 & 0.14 & 0.09 & 0.03 \\
 & Stacked Windows A and B & 0.22 & 0.38 & 0.34 & 0.04 & 0.04 \\
 & Mean of Windows A and B & 0.15 & 0.58 & 0.17 & 0.05 & 0.02 \\
 & Window A only & 0.10 & 0.80 & 0.10 & 0.06 & 0.02 \\
 & Window B only & 0.29 & 0.66 & 0.35 & 0.10 & 0.08 \\ \midrule
\multirow{5}{*}{Rwanda} & Stacked Windows A and B (RGB only) & - & - & 0.61 & - & 0.26 \\
 & Stacked Windows A and B & - & - & 0.58 & - & 0.27 \\
 & Mean of Windows A and B & - & - & 0.57 & - & 0.26 \\
 & Window A only & - & - & 0.50 & - & 0.23 \\
 & Window B only & - & - & 0.64 & - & 0.34 \\ 
 \bottomrule
\end{tabular}%
}
\end{table*}

\begin{table*}[ht]
\caption{Performance for input channel ablations in all test countries (names starting with S-Z): multispectral (RGB-NIR vs. RGB only) and multi-temporal (Window A and Window B) input channels.}
\label{tab:exp1-channels-fullc}
\resizebox{\textwidth}{!}{%
\begin{tabular}{@{}ccccccc@{}}
\toprule
\textbf{Test country} & \textbf{Channels} & \textbf{Pixel IoU} & \textbf{Pixel precision} & \textbf{Pixel recall} & \textbf{Object precision} & \textbf{Object recall} \\ \midrule
\multirow{5}{*}{Slovakia} & Stacked Windows A and B (RGB only) & 0.91 & 0.97 & 0.94 & 0.47 & 0.53 \\
 & Stacked Windows A and B & 0.92 & 0.98 & 0.95 & 0.52 & 0.55 \\
 & Mean of Windows A and B & 0.92 & 0.97 & 0.94 & 0.49 & 0.54 \\
 & Window A only & 0.91 & 0.97 & 0.94 & 0.46 & 0.52 \\
 & Window B only & 0.91 & 0.97 & 0.94 & 0.48 & 0.54 \\ \midrule
\multirow{5}{*}{Slovenia} & Stacked Windows A and B (RGB only) & 0.58 & 0.90 & 0.62 & 0.27 & 0.18 \\
 & Stacked Windows A and B & 0.58 & 0.91 & 0.61 & 0.30 & 0.18 \\
 & Mean of Windows A and B & 0.54 & 0.88 & 0.59 & 0.27 & 0.16 \\
 & Window A only & 0.55 & 0.88 & 0.59 & 0.27 & 0.17 \\
 & Window B only & 0.52 & 0.87 & 0.57 & 0.24 & 0.15 \\ \midrule
\multirow{5}{*}{South Africa} & Stacked Windows A and B (RGB only) & 0.79 & 0.89 & 0.88 & 0.53 & 0.54 \\
 & Stacked Windows A and B & 0.80 & 0.90 & 0.87 & 0.55 & 0.54 \\
 & Mean of Windows A and B & 0.78 & 0.88 & 0.88 & 0.49 & 0.53 \\
 & Window A only & 0.78 & 0.88 & 0.87 & 0.47 & 0.52 \\
 & Window B only & 0.79 & 0.89 & 0.87 & 0.52 & 0.53 \\ \midrule
\multirow{5}{*}{Spain} & Stacked Windows A and B (RGB only) & 0.73 & 0.96 & 0.75 & 0.34 & 0.19 \\
 & Stacked Windows A and B & 0.73 & 0.96 & 0.75 & 0.34 & 0.19 \\
 & Mean of Windows A and B & 0.72 & 0.96 & 0.74 & 0.32 & 0.18 \\
 & Window A only & 0.71 & 0.96 & 0.74 & 0.31 & 0.17 \\
 & Window B only & 0.71 & 0.96 & 0.73 & 0.32 & 0.18 \\ \midrule
\multirow{5}{*}{Sweden} & Stacked Windows A and B (RGB only) & 0.80 & 0.94 & 0.85 & 0.40 & 0.50 \\
 & Stacked Windows A and B & 0.81 & 0.94 & 0.85 & 0.42 & 0.51 \\
 & Mean of Windows A and B & 0.80 & 0.93 & 0.84 & 0.40 & 0.48 \\
 & Window A only & 0.81 & 0.94 & 0.85 & 0.39 & 0.49 \\
 & Window B only & 0.79 & 0.92 & 0.84 & 0.38 & 0.47 \\ \midrule
\multirow{5}{*}{Vietnam} & Stacked Windows A and B (RGB only) & 0.36 & 0.90 & 0.37 & 0.11 & 0.08 \\
 & Stacked Windows A and B & 0.45 & 0.89 & 0.47 & 0.17 & 0.12 \\
 & Mean of Windows A and B & 0.33 & 0.88 & 0.34 & 0.09 & 0.06 \\
 & Window A only & 0.35 & 0.87 & 0.37 & 0.09 & 0.06 \\
 & Window B only & 0.39 & 0.90 & 0.41 & 0.14 & 0.10 \\ 
 \bottomrule
\end{tabular}%
}
\end{table*}

\subsubsection{Multiple random seeds}

The results in Tables 3, 4, and 5 of the main paper were run with one arbitrarily-chosen random seed. In supplement Tables~\ref{tab:exp1-masktype-seeds}-\ref{tab:exp1-masktype-seedsb}, we report the average results across three random seeds for all test countries for the mask type experiment (Table 3 in the main paper). The standard deviation across random seeds for each experiment and test country are very small (0 or close to 0 for most metrics).

\subsubsection{Benchmarking example}
We suggest benchmarking performance on the per-country test sets and reporting individual country results, the mean across all countries, or the minimum across countries (worst-case performance). Supplement Table \ref{tab:benchmarking-example} reports these metrics for the best model evaluated in this paper (U-net with EfficientNet-b3 backbone). 

\begin{table*}[ht]
\caption{Performance metrics for U-net with EfficientNet-b3 backbone with 3-class masks, ignoring background pixels for presence-only samples. We only report recall metrics for presence-only countries.}
\label{tab:benchmarking-example}
\centering
\begin{tabular}{@{}cccccc@{}}
\toprule
\textbf{Test country} & \textbf{Pixel IoU} & \textbf{Pixel precision} & \textbf{Pixel recall} & \textbf{Object precision} & \textbf{Object recall} \\ \midrule
Austria & 0.70 & 0.90 & 0.76 & 0.44 & 0.39 \\ 
Belgium & 0.75 & 0.92 & 0.80 & 0.57 & 0.58 \\
Brazil & - & - & 0.96 & - & 0.58 \\
Cambodia & 0.43 & 0.95 & 0.44 & 0.26 & 0.20 \\ 
Corsica & 0.48 & 0.79 & 0.55 & 0.21 & 0.17 \\ 
Croatia & 0.68 & 0.89 & 0.74 & 0.25 & 0.34 \\ 
Denmark & 0.83 & 0.93 & 0.88 & 0.45 & 0.60 \\
Estonia & 0.79 & 0.91 & 0.86 & 0.47 & 0.43 \\ 
Finland & 0.83 & 0.96 & 0.87 & 0.55 & 0.57 \\ 
France & 0.79 & 0.89 & 0.88 & 0.55 & 0.58 \\ 
Germany & 0.79 & 0.87 & 0.90 & 0.43 & 0.42 \\ 
India & - & - & 0.22 & - & 0.06 \\ 
Kenya & - & - & 0.49 & - & 0.10 \\ 
Latvia & 0.81 & 0.94 & 0.86 & 0.44 & 0.45 \\ 
Lithuania & 0.74 & 0.88 & 0.83 & 0.37 & 0.41 \\ 
Luxembourg & 0.79 & 0.96 & 0.82 & 0.47 & 0.51 \\
Netherlands & 0.75 & 0.92 & 0.80 & 0.53 & 0.45 \\ 
Portugal & 0.12 & 0.67 & 0.12 & 0.07 & 0.03 \\ 
Rwanda & - & - & 0.57 & - & 0.30 \\ 
Slovakia & 0.92 & 0.98 & 0.95 & 0.50 & 0.55 \\
Slovenia & 0.59 & 0.90 & 0.63 & 0.33 & 0.20 \\ 
South Africa & 0.79 & 0.89 & 0.88 & 0.51 & 0.55 \\ 
Spain & 0.74 & 0.96 & 0.76 & 0.36 & 0.20 \\ 
Sweden & 0.81 & 0.94 & 0.85 & 0.40 & 0.51 \\ 
Vietnam & 0.47 & 0.89 & 0.49 & 0.18 & 0.13 \\ \midrule
Mean & 0.70 & 0.90 & 0.72 & 0.40 & 0.37 \\
Minimum & 0.12 & 0.67 & 0.12 & 0.07 & 0.03 \\
\bottomrule
\end{tabular}%
\end{table*}

\begin{table*}[ht]
\caption{Performance metrics for different target mask formats in all test countries averaged over 3 random seeds (names starting with A-K). We only report recall metrics for presence-only countries.}
\label{tab:exp1-masktype-seeds}
\resizebox{\textwidth}{!}{%
\begin{tabular}{@{}ccccccl@{}}
\toprule
\textbf{Test country} & \textbf{Mask type} & \textbf{Pixel IoU} & \textbf{Pixel precision} & \textbf{Pixel recall} & \textbf{Object precision} & \textbf{Object recall} \\ \midrule
\multirow{4}{*}{Austria} & 2-class & 0.77 $\pm$ 0.00 & 0.83 $\pm$ 0.00 & 0.91 $\pm$ 0.00 & 0.38 $\pm$ 0.01 & 0.11 $\pm$ 0.00 \\
 & 2-class (ignore presence-only bg) & 0.77 $\pm$ 0.00 & 0.83 $\pm$ 0.00 & 0.91 $\pm$ 0.00 & 0.37 $\pm$ 0.02 & 0.12 $\pm$ 0.00 \\
 & 3-class & 0.71 $\pm$ 0.00 & 0.90 $\pm$ 0.00 & 0.77 $\pm$ 0.00 & 0.45 $\pm$ 0.00 & 0.39 $\pm$ 0.00 \\
 & 3-class (ignore presence-only bg) & 0.71 $\pm$ 0.00 & 0.90 $\pm$ 0.01 & 0.76 $\pm$ 0.00 & 0.45 $\pm$ 0.00 & 0.39 $\pm$ 0.00 \\ \midrule
\multirow{4}{*}{Belgium} & 2-class & 0.80 $\pm$ 0.00 & 0.85 $\pm$ 0.00 & 0.92 $\pm$ 0.00 & 0.48 $\pm$ 0.02 & 0.23 $\pm$ 0.00 \\
 & 2-class (ignore presence-only bg) & 0.80 $\pm$ 0.00 & 0.86 $\pm$ 0.00 & 0.92 $\pm$ 0.01 & 0.48 $\pm$ 0.01 & 0.23 $\pm$ 0.00 \\
 & 3-class & 0.75 $\pm$ 0.00 & 0.93 $\pm$ 0.00 & 0.80 $\pm$ 0.00 & 0.57 $\pm$ 0.01 & 0.58 $\pm$ 0.00 \\
 & 3-class (ignore presence-only bg) & 0.75 $\pm$ 0.00 & 0.93 $\pm$ 0.00 & 0.80 $\pm$ 0.00 & 0.57 $\pm$ 0.01 & 0.58 $\pm$ 0.00 \\ \midrule
\multirow{4}{*}{Brazil} & 2-class & - & - & 1.00 $\pm$ 0.00 & - & 0.12 $\pm$ 0.02 \\
 & 2-class (ignore presence-only bg) & - & - & 1.00 $\pm$ 0.00 & - & 0.13 $\pm$ 0.01 \\
 & 3-class & - & - & 0.96 $\pm$ 0.00 & - & 0.58 $\pm$ 0.01 \\
 & 3-class (ignore presence-only bg) & - & - & 0.97 $\pm$ 0.00 & - & 0.58 $\pm$ 0.01 \\ \midrule
\multirow{4}{*}{Cambodia} & 2-class & 0.76 $\pm$ 0.00 & 0.78 $\pm$ 0.00 & 0.97 $\pm$ 0.00 & 0.04 $\pm$ 0.01 & 0.00 $\pm$ 0.00 \\
 & 2-class (ignore presence-only bg) & 0.76 $\pm$ 0.00 & 0.78 $\pm$ 0.00 & 0.97 $\pm$ 0.00 & 0.04 $\pm$ 0.02 & 0.00 $\pm$ 0.00 \\
 & 3-class & 0.41 $\pm$ 0.02 & 0.95 $\pm$ 0.00 & 0.42 $\pm$ 0.02 & 0.23 $\pm$ 0.01 & 0.18 $\pm$ 0.01 \\
 & 3-class (ignore presence-only bg) & 0.40 $\pm$ 0.01 & 0.95 $\pm$ 0.00 & 0.41 $\pm$ 0.01 & 0.22 $\pm$ 0.01 & 0.17 $\pm$ 0.01 \\ \midrule
\multirow{4}{*}{Corsica} & 2-class & 0.49 $\pm$ 0.01 & 0.68 $\pm$ 0.03 & 0.64 $\pm$ 0.01 & 0.18 $\pm$ 0.01 & 0.08 $\pm$ 0.00 \\
 & 2-class (ignore presence-only bg) & 0.48 $\pm$ 0.01 & 0.67 $\pm$ 0.01 & 0.62 $\pm$ 0.01 & 0.19 $\pm$ 0.01 & 0.09 $\pm$ 0.00 \\
 & 3-class & 0.48 $\pm$ 0.01 & 0.79 $\pm$ 0.01 & 0.54 $\pm$ 0.02 & 0.24 $\pm$ 0.01 & 0.17 $\pm$ 0.01 \\
 & 3-class (ignore presence-only bg) & 0.46 $\pm$ 0.01 & 0.79 $\pm$ 0.01 & 0.53 $\pm$ 0.01 & 0.24 $\pm$ 0.01 & 0.18 $\pm$ 0.00 \\ \midrule
\multirow{4}{*}{Croatia} & 2-class & 0.76 $\pm$ 0.00 & 0.80 $\pm$ 0.00 & 0.94 $\pm$ 0.00 & 0.26 $\pm$ 0.01 & 0.08 $\pm$ 0.00 \\
 & 2-class (ignore presence-only bg) & 0.76 $\pm$ 0.00 & 0.80 $\pm$ 0.01 & 0.93 $\pm$ 0.00 & 0.26 $\pm$ 0.01 & 0.09 $\pm$ 0.00 \\
 & 3-class & 0.68 $\pm$ 0.00 & 0.89 $\pm$ 0.00 & 0.74 $\pm$ 0.00 & 0.25 $\pm$ 0.01 & 0.34 $\pm$ 0.01 \\
 & 3-class (ignore presence-only bg) & 0.67 $\pm$ 0.00 & 0.89 $\pm$ 0.00 & 0.74 $\pm$ 0.00 & 0.25 $\pm$ 0.00 & 0.33 $\pm$ 0.01 \\ \midrule
\multirow{4}{*}{Denmark} & 2-class & 0.84 $\pm$ 0.00 & 0.89 $\pm$ 0.00 & 0.94 $\pm$ 0.00 & 0.40 $\pm$ 0.02 & 0.25 $\pm$ 0.00 \\
 & 2-class (ignore presence-only bg) & 0.84 $\pm$ 0.00 & 0.89 $\pm$ 0.00 & 0.94 $\pm$ 0.00 & 0.41 $\pm$ 0.03 & 0.27 $\pm$ 0.01 \\
 & 3-class & 0.83 $\pm$ 0.00 & 0.93 $\pm$ 0.00 & 0.88 $\pm$ 0.00 & 0.46 $\pm$ 0.01 & 0.60 $\pm$ 0.00 \\
 & 3-class (ignore presence-only bg) & 0.83 $\pm$ 0.00 & 0.93 $\pm$ 0.00 & 0.88 $\pm$ 0.00 & 0.46 $\pm$ 0.01 & 0.60 $\pm$ 0.01 \\ \midrule
\multirow{4}{*}{Estonia} & 2-class & 0.81 $\pm$ 0.00 & 0.88 $\pm$ 0.00 & 0.92 $\pm$ 0.00 & 0.49 $\pm$ 0.01 & 0.29 $\pm$ 0.00 \\
 & 2-class (ignore presence-only bg) & 0.81 $\pm$ 0.01 & 0.88 $\pm$ 0.00 & 0.92 $\pm$ 0.00 & 0.50 $\pm$ 0.02 & 0.30 $\pm$ 0.00 \\
 & 3-class & 0.79 $\pm$ 0.00 & 0.92 $\pm$ 0.00 & 0.85 $\pm$ 0.01 & 0.47 $\pm$ 0.02 & 0.42 $\pm$ 0.00 \\
 & 3-class (ignore presence-only bg) & 0.79 $\pm$ 0.00 & 0.91 $\pm$ 0.00 & 0.86 $\pm$ 0.00 & 0.47 $\pm$ 0.01 & 0.42 $\pm$ 0.00 \\ \midrule
\multirow{4}{*}{Finland} & 2-class & 0.87 $\pm$ 0.00 & 0.90 $\pm$ 0.00 & 0.96 $\pm$ 0.00 & 0.44 $\pm$ 0.01 & 0.18 $\pm$ 0.00 \\
 & 2-class (ignore presence-only bg) & 0.87 $\pm$ 0.00 & 0.90 $\pm$ 0.00 & 0.96 $\pm$ 0.00 & 0.44 $\pm$ 0.02 & 0.18 $\pm$ 0.01 \\
 & 3-class & 0.83 $\pm$ 0.00 & 0.96 $\pm$ 0.00 & 0.86 $\pm$ 0.00 & 0.55 $\pm$ 0.00 & 0.56 $\pm$ 0.01 \\
 & 3-class (ignore presence-only bg) & 0.83 $\pm$ 0.00 & 0.96 $\pm$ 0.00 & 0.86 $\pm$ 0.00 & 0.54 $\pm$ 0.01 & 0.56 $\pm$ 0.00 \\ \midrule
\multirow{4}{*}{France} & 2-class & 0.81 $\pm$ 0.00 & 0.85 $\pm$ 0.00 & 0.95 $\pm$ 0.00 & 0.38 $\pm$ 0.01 & 0.15 $\pm$ 0.01 \\
 & 2-class (ignore presence-only bg) & 0.82 $\pm$ 0.00 & 0.86 $\pm$ 0.00 & 0.95 $\pm$ 0.00 & 0.38 $\pm$ 0.01 & 0.15 $\pm$ 0.00 \\
 & 3-class & 0.79 $\pm$ 0.00 & 0.89 $\pm$ 0.00 & 0.87 $\pm$ 0.00 & 0.54 $\pm$ 0.01 & 0.58 $\pm$ 0.01 \\
 & 3-class (ignore presence-only bg) & 0.79 $\pm$ 0.00 & 0.89 $\pm$ 0.00 & 0.87 $\pm$ 0.00 & 0.54 $\pm$ 0.01 & 0.57 $\pm$ 0.00 \\ \midrule
\multirow{4}{*}{Germany} & 2-class & 0.80 $\pm$ 0.00 & 0.84 $\pm$ 0.00 & 0.94 $\pm$ 0.01 & 0.40 $\pm$ 0.03 & 0.18 $\pm$ 0.01 \\
 & 2-class (ignore presence-only bg) & 0.80 $\pm$ 0.00 & 0.84 $\pm$ 0.00 & 0.94 $\pm$ 0.00 & 0.40 $\pm$ 0.03 & 0.19 $\pm$ 0.00 \\
 & 3-class & 0.79 $\pm$ 0.00 & 0.87 $\pm$ 0.00 & 0.89 $\pm$ 0.01 & 0.40 $\pm$ 0.02 & 0.40 $\pm$ 0.01 \\
 & 3-class (ignore presence-only bg) & 0.79 $\pm$ 0.00 & 0.87 $\pm$ 0.00 & 0.89 $\pm$ 0.00 & 0.41 $\pm$ 0.01 & 0.40 $\pm$ 0.01 \\ \midrule
\multirow{4}{*}{India} & 2-class & - & - & 0.99 $\pm$ 0.00 & - & 0.00 $\pm$ 0.00 \\
 & 2-class (ignore presence-only bg) & - & - & 0.99 $\pm$ 0.00 & - & 0.00 $\pm$ 0.00 \\
 & 3-class & - & - & 0.22 $\pm$ 0.03 & - & 0.05 $\pm$ 0.01 \\
 & 3-class (ignore presence-only bg) & - & - & 0.22 $\pm$ 0.02 & - & 0.05 $\pm$ 0.01 \\ \midrule
\multirow{4}{*}{Kenya} & 2-class & - & - & 0.96 $\pm$ 0.01 & - & 0.00 $\pm$ 0.00 \\
 & 2-class (ignore presence-only bg) & - & - & 0.96 $\pm$ 0.01 & - & 0.00 $\pm$ 0.01 \\
 & 3-class & - & - & 0.50 $\pm$ 0.02 & - & 0.09 $\pm$ 0.02 \\
 & 3-class (ignore presence-only bg) & - & - & 0.47 $\pm$ 0.02 & - & 0.08 $\pm$ 0.01 \\ 
 \bottomrule
\end{tabular}%
}
\end{table*}

\begin{table*}[ht]
\caption{Performance metrics for different target mask formats in all test countries averaged over 3 random seeds (names starting with L-Z). We only report recall metrics for presence-only countries.}
\label{tab:exp1-masktype-seedsb}
\resizebox{\textwidth}{!}{%
\begin{tabular}{@{}ccccccl@{}}
\toprule
\textbf{Test country} & \textbf{Mask type} & \textbf{Pixel IoU} & \textbf{Pixel precision} & \textbf{Pixel recall} & \textbf{Object precision} & \textbf{Object recall} \\ \midrule
\multirow{4}{*}{Latvia} & 2-class & 0.83 $\pm$ 0.00 & 0.90 $\pm$ 0.00 & 0.92 $\pm$ 0.00 & 0.44 $\pm$ 0.00 & 0.27 $\pm$ 0.00 \\
 & 2-class (ignore presence-only bg) & 0.84 $\pm$ 0.00 & 0.90 $\pm$ 0.00 & 0.92 $\pm$ 0.00 & 0.45 $\pm$ 0.01 & 0.27 $\pm$ 0.01 \\
 & 3-class & 0.81 $\pm$ 0.00 & 0.94 $\pm$ 0.00 & 0.85 $\pm$ 0.00 & 0.45 $\pm$ 0.01 & 0.45 $\pm$ 0.00 \\
 & 3-class (ignore presence-only bg) & 0.81 $\pm$ 0.00 & 0.94 $\pm$ 0.00 & 0.86 $\pm$ 0.00 & 0.45 $\pm$ 0.01 & 0.45 $\pm$ 0.00 \\ \midrule
\multirow{4}{*}{Lithuania} & 2-class & 0.77 $\pm$ 0.00 & 0.82 $\pm$ 0.00 & 0.93 $\pm$ 0.00 & 0.39 $\pm$ 0.02 & 0.19 $\pm$ 0.00 \\
 & 2-class (ignore presence-only bg) & 0.77 $\pm$ 0.00 & 0.82 $\pm$ 0.01 & 0.93 $\pm$ 0.00 & 0.39 $\pm$ 0.01 & 0.19 $\pm$ 0.00 \\
 & 3-class & 0.74 $\pm$ 0.01 & 0.88 $\pm$ 0.01 & 0.83 $\pm$ 0.01 & 0.38 $\pm$ 0.01 & 0.42 $\pm$ 0.00 \\
 & 3-class (ignore presence-only bg) & 0.74 $\pm$ 0.00 & 0.88 $\pm$ 0.00 & 0.82 $\pm$ 0.00 & 0.38 $\pm$ 0.01 & 0.41 $\pm$ 0.00 \\ \midrule
\multirow{4}{*}{Luxembourg} & 2-class & 0.86 $\pm$ 0.00 & 0.88 $\pm$ 0.00 & 0.97 $\pm$ 0.00 & 0.24 $\pm$ 0.00 & 0.05 $\pm$ 0.00 \\
 & 2-class (ignore presence-only bg) & 0.86 $\pm$ 0.00 & 0.88 $\pm$ 0.00 & 0.97 $\pm$ 0.00 & 0.25 $\pm$ 0.03 & 0.05 $\pm$ 0.01 \\
 & 3-class & 0.79 $\pm$ 0.00 & 0.96 $\pm$ 0.00 & 0.81 $\pm$ 0.00 & 0.45 $\pm$ 0.01 & 0.51 $\pm$ 0.00 \\
 & 3-class (ignore presence-only bg) & 0.79 $\pm$ 0.00 & 0.97 $\pm$ 0.00 & 0.81 $\pm$ 0.00 & 0.46 $\pm$ 0.01 & 0.51 $\pm$ 0.00 \\ \midrule
\multirow{4}{*}{Netherlands} & 2-class & 0.79 $\pm$ 0.01 & 0.86 $\pm$ 0.01 & 0.91 $\pm$ 0.00 & 0.51 $\pm$ 0.01 & 0.24 $\pm$ 0.00 \\
 & 2-class (ignore presence-only bg) & 0.79 $\pm$ 0.01 & 0.85 $\pm$ 0.00 & 0.91 $\pm$ 0.01 & 0.50 $\pm$ 0.01 & 0.24 $\pm$ 0.00 \\
 & 3-class & 0.74 $\pm$ 0.00 & 0.92 $\pm$ 0.00 & 0.80 $\pm$ 0.00 & 0.52 $\pm$ 0.01 & 0.45 $\pm$ 0.00 \\
 & 3-class (ignore presence-only bg) & 0.75 $\pm$ 0.00 & 0.92 $\pm$ 0.00 & 0.80 $\pm$ 0.00 & 0.52 $\pm$ 0.01 & 0.45 $\pm$ 0.00 \\ \midrule
\multirow{4}{*}{Portugal} & 2-class & 0.31 $\pm$ 0.03 & 0.62 $\pm$ 0.03 & 0.39 $\pm$ 0.06 & 0.06 $\pm$ 0.01 & 0.01 $\pm$ 0.00 \\
 & 2-class (ignore presence-only bg) & 0.33 $\pm$ 0.04 & 0.54 $\pm$ 0.09 & 0.49 $\pm$ 0.12 & 0.05 $\pm$ 0.01 & 0.01 $\pm$ 0.00 \\
 & 3-class & 0.17 $\pm$ 0.06 & 0.66 $\pm$ 0.27 & 0.24 $\pm$ 0.16 & 0.07 $\pm$ 0.02 & 0.04 $\pm$ 0.01 \\
 & 3-class (ignore presence-only bg) & 0.14 $\pm$ 0.02 & 0.76 $\pm$ 0.11 & 0.15 $\pm$ 0.03 & 0.08 $\pm$ 0.03 & 0.03 $\pm$ 0.01 \\ \midrule
\multirow{4}{*}{Rwanda} & 2-class & - & - & 0.99 $\pm$ 0.00 & - & 0.00 $\pm$ 0.00 \\
 & 2-class (ignore presence-only bg) & - & - & 0.99 $\pm$ 0.00 & - & 0.00 $\pm$ 0.00 \\
 & 3-class & - & - & 0.60 $\pm$ 0.02 & - & 0.27 $\pm$ 0.04 \\
 & 3-class (ignore presence-only bg) & - & - & 0.59 $\pm$ 0.02 & - & 0.27 $\pm$ 0.02 \\ \midrule
\multirow{4}{*}{Slovakia} & 2-class & 0.93 $\pm$ 0.00 & 0.95 $\pm$ 0.00 & 0.98 $\pm$ 0.00 & 0.60 $\pm$ 0.01 & 0.40 $\pm$ 0.01 \\
 & 2-class (ignore presence-only bg) & 0.93 $\pm$ 0.00 & 0.95 $\pm$ 0.00 & 0.98 $\pm$ 0.00 & 0.60 $\pm$ 0.03 & 0.41 $\pm$ 0.01 \\
 & 3-class & 0.92 $\pm$ 0.00 & 0.98 $\pm$ 0.00 & 0.95 $\pm$ 0.00 & 0.51 $\pm$ 0.01 & 0.55 $\pm$ 0.00 \\
 & 3-class (ignore presence-only bg) & 0.92 $\pm$ 0.00 & 0.98 $\pm$ 0.00 & 0.94 $\pm$ 0.00 & 0.52 $\pm$ 0.01 & 0.55 $\pm$ 0.00 \\ \midrule
\multirow{4}{*}{Slovenia} & 2-class & 0.69 $\pm$ 0.01 & 0.79 $\pm$ 0.00 & 0.84 $\pm$ 0.01 & 0.30 $\pm$ 0.02 & 0.08 $\pm$ 0.00 \\
 & 2-class (ignore presence-only bg) & 0.69 $\pm$ 0.00 & 0.79 $\pm$ 0.01 & 0.85 $\pm$ 0.01 & 0.31 $\pm$ 0.02 & 0.09 $\pm$ 0.00 \\
 & 3-class & 0.59 $\pm$ 0.01 & 0.90 $\pm$ 0.00 & 0.63 $\pm$ 0.01 & 0.30 $\pm$ 0.00 & 0.19 $\pm$ 0.00 \\
 & 3-class (ignore presence-only bg) & 0.59 $\pm$ 0.01 & 0.91 $\pm$ 0.00 & 0.63 $\pm$ 0.01 & 0.31 $\pm$ 0.01 & 0.19 $\pm$ 0.01 \\ \midrule
\multirow{4}{*}{South Africa} & 2-class & 0.82 $\pm$ 0.00 & 0.85 $\pm$ 0.01 & 0.96 $\pm$ 0.00 & 0.44 $\pm$ 0.01 & 0.2 $\pm$ 0.01 \\
 & 2-class (ignore presence-only bg) & 0.82 $\pm$ 0.00 & 0.85 $\pm$ 0.00 & 0.96 $\pm$ 0.01 & 0.43 $\pm$ 0.01 & 0.2 $\pm$ 0.01 \\
 & 3-class & 0.80 $\pm$ 0.00 & 0.89 $\pm$ 0.00 & 0.88 $\pm$ 0.00 & 0.53 $\pm$ 0.02 & 0.54 $\pm$ 0.00 \\
 & 3-class (ignore presence-only bg) & 0.80 $\pm$ 0.00 & 0.89 $\pm$ 0.00 & 0.88 $\pm$ 0.00 & 0.52 $\pm$ 0.02 & 0.55 $\pm$ 0.01 \\ \midrule
\multirow{4}{*}{Spain} & 2-class & 0.84 $\pm$ 0.00 & 0.87 $\pm$ 0.00 & 0.95 $\pm$ 0.00 & 0.34 $\pm$ 0.01 & 0.05 $\pm$ 0.00 \\
 & 2-class (ignore presence-only bg) & 0.84 $\pm$ 0.00 & 0.88 $\pm$ 0.01 & 0.95 $\pm$ 0.00 & 0.34 $\pm$ 0.03 & 0.05 $\pm$ 0.00 \\
 & 3-class & 0.74 $\pm$ 0.00 & 0.96 $\pm$ 0.00 & 0.76 $\pm$ 0.00 & 0.34 $\pm$ 0.01 & 0.2 $\pm$ 0.00 \\
 & 3-class (ignore presence-only bg) & 0.73 $\pm$ 0.00 & 0.96 $\pm$ 0.00 & 0.75 $\pm$ 0.00 & 0.34 $\pm$ 0.01 & 0.19 $\pm$ 0.00 \\ \midrule
\multirow{4}{*}{Sweden} & 2-class & 0.83 $\pm$ 0.00 & 0.88 $\pm$ 0.00 & 0.93 $\pm$ 0.00 & 0.35 $\pm$ 0.01 & 0.19 $\pm$ 0.01 \\
 & 2-class (ignore presence-only bg) & 0.83 $\pm$ 0.00 & 0.89 $\pm$ 0.01 & 0.93 $\pm$ 0.00 & 0.36 $\pm$ 0.04 & 0.2 $\pm$ 0.01 \\
 & 3-class & 0.81 $\pm$ 0.00 & 0.94 $\pm$ 0.00 & 0.86 $\pm$ 0.00 & 0.41 $\pm$ 0.01 & 0.51 $\pm$ 0.00 \\
 & 3-class (ignore presence-only bg) & 0.81 $\pm$ 0.00 & 0.94 $\pm$ 0.00 & 0.85 $\pm$ 0.00 & 0.41 $\pm$ 0.01 & 0.51 $\pm$ 0.01 \\ \midrule
\multirow{4}{*}{Vietnam} & 2-class & 0.67 $\pm$ 0.00 & 0.7 $\pm$ 0.00 & 0.94 $\pm$ 0.00 & 0.1 $\pm$ 0.01 & 0.01 $\pm$ 0.00 \\
 & 2-class (ignore presence-only bg) & 0.67 $\pm$ 0.00 & 0.7 $\pm$ 0.00 & 0.94 $\pm$ 0.00 & 0.11 $\pm$ 0.01 & 0.01 $\pm$ 0.00 \\
 & 3-class & 0.46 $\pm$ 0.02 & 0.89 $\pm$ 0.00 & 0.49 $\pm$ 0.02 & 0.17 $\pm$ 0.01 & 0.12 $\pm$ 0.01 \\
 & 3-class (ignore presence-only bg) & 0.47 $\pm$ 0.01 & 0.89 $\pm$ 0.01 & 0.50 $\pm$ 0.01 & 0.19 $\pm$ 0.01 & 0.12 $\pm$ 0.00 \\ \bottomrule
\end{tabular}%
}
\end{table*}

\end{document}